\documentclass[runningheads]{llncs}

\usepackage[mobile]{eccv}
\usepackage{eccvabbrv}
\usepackage{graphicx}
\usepackage{booktabs}
\usepackage{pifont}
\usepackage[accsupp]{axessibility} 
\usepackage{wrapfig}
\usepackage{algorithm,algpseudocode}

\graphicspath{{sections/image/}} 

\usepackage{amsmath,amsfonts,bm}

\def\eqref#1{equation~\ref{#1}}

\def\1{\bm{1}}

\DeclareMathAlphabet{\mathsfit}{\encodingdefault}{\sfdefault}{m}{sl}
\SetMathAlphabet{\mathsfit}{bold}{\encodingdefault}{\sfdefault}{bx}{n}

\newcommand{\methodName}{ActionPlan}

\algrenewcommand\algorithmicrequire{\textbf{Input:}}
\algrenewcommand\algorithmicensure{\textbf{Output:}}

\AtEndPreamble{
    \usepackage[capitalize]{cleveref}
    \crefname{section}{Sec.}{Secs.}
    \Crefname{section}{Section}{Sections}
    \Crefname{table}{Table}{Tables}
    \crefname{table}{Tab.}{Tabs.}
}     

\usepackage[pagebackref,breaklinks,colorlinks,citecolor=eccvblue]{hyperref}

\usepackage{orcidlink}
\usepackage{multirow}
\usepackage{pifont}  
\usepackage{wrapfig}  
\usepackage{adjustbox}
\usepackage{float}
\usepackage{amsmath}

\usepackage{makecell}
\usepackage{array}
\usepackage{graphicx}

\begin{document}

\title{\methodName{}: Future-Aware Streaming Motion Synthesis via Frame-Level Action Planning} 

\titlerunning{Future-Aware Streaming Motion Synthesis}

\author{
  Eric Nazarenus$^{*}$\inst{1}\orcidlink{...} \and
  Chuqiao Li$^{*\dagger}$\inst{1}\orcidlink{...} \and
  Yannan He\inst{1}\orcidlink{...} \and
  Xianghui Xie\inst{1,2}\orcidlink{...} \and
  Jan Eric Lenssen\inst{2}\orcidlink{...} \and
  Gerard Pons-Moll\inst{1,2}\orcidlink{...}
}

\authorrunning{E.~Nazarenus and C.~Li et al.}

\institute{Tübingen AI Center, University of Tübingen, Germany \and
Max Planck Institute for Informatics, Saarland Informatics Campus, Germany\\
\url{https://coral79.github.io/ActionPlan/} }

\maketitle

\begingroup
\renewcommand{\thefootnote}{}
\phantomsection
\footnotetext{$^{*}$ Equal contribution. \quad $^{\dagger}$ Corresponding author}
\endgroup

\begin{figure}[ht]
  \centering
  \vspace{-0.8cm}
  \includegraphics[width=\textwidth]{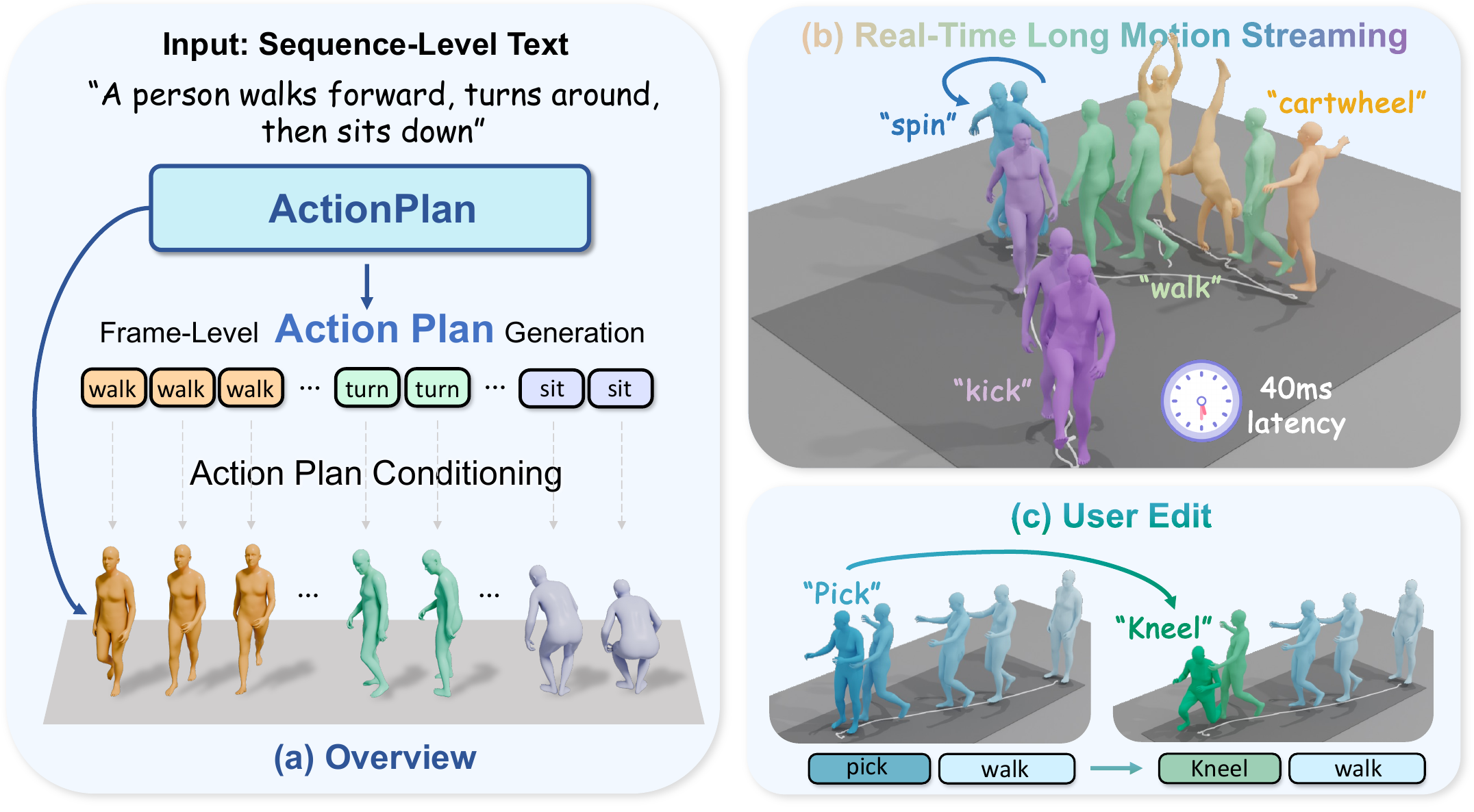}
  \vspace{-0.7cm}
  \caption{\methodName{} decouples high-level action planning from low-level motion generation in a single generative model (a). By conditioning motion synthesis on generated action plans, \methodName{} achieves online generation (b) without the typical accuracy drop that happens in existing streaming methods and supports localized edits (c).}
  \label{fig:teaser_figure}
  \vspace{-1.0cm}
\end{figure}

\begin{abstract}
  We present \textbf{\methodName{}}, a unified motion diffusion framework that bridges real-time streaming with high-quality offline generation within a single model. The core idea is to introduce a per-frame action plan: the model predicts frame-level text latents that act as dense semantic anchors throughout denoising, and uses them to denoise the full motion sequence with combined semantic and motion cues. To support this structured workflow, we design latent-specific diffusion steps, allowing each motion latent to be denoised independently and sampled in flexible orders at inference. As a result, \methodName\ can run in a history-conditioned, future-aware mode for real-time streaming, while also supporting high-quality offline generation. The same mechanism further enables zero-shot motion editing and in-betweening without additional models. Experiments demonstrate that our real-time streaming is $5.25\times$ faster while also achieving 18\% motion quality improvement over the best previous method in terms of FID.
\end{abstract}
\section{Introduction}
\label{sec:intro}

Text-driven human motion generation~\cite{tevet2022human,Guo2022CVPR_humanml3d, zhang2023generating, zhang2024motiondiffuse} has emerged as a key technology for enabling realistic interactions in digital humans, virtual reality, and autonomous agents. An ideal motion synthesis framework should simultaneously satisfy three core requirements: high \textit{semantic fidelity} to complex textual prompts, \textit{on-the-fly streaming} for low-latency interaction, and sufficient \textit{versatility} to support different tasks such as zero-shot editing and in-betweening.

\begin{figure}[t]
  \centering
  \includegraphics[width=0.95\textwidth]{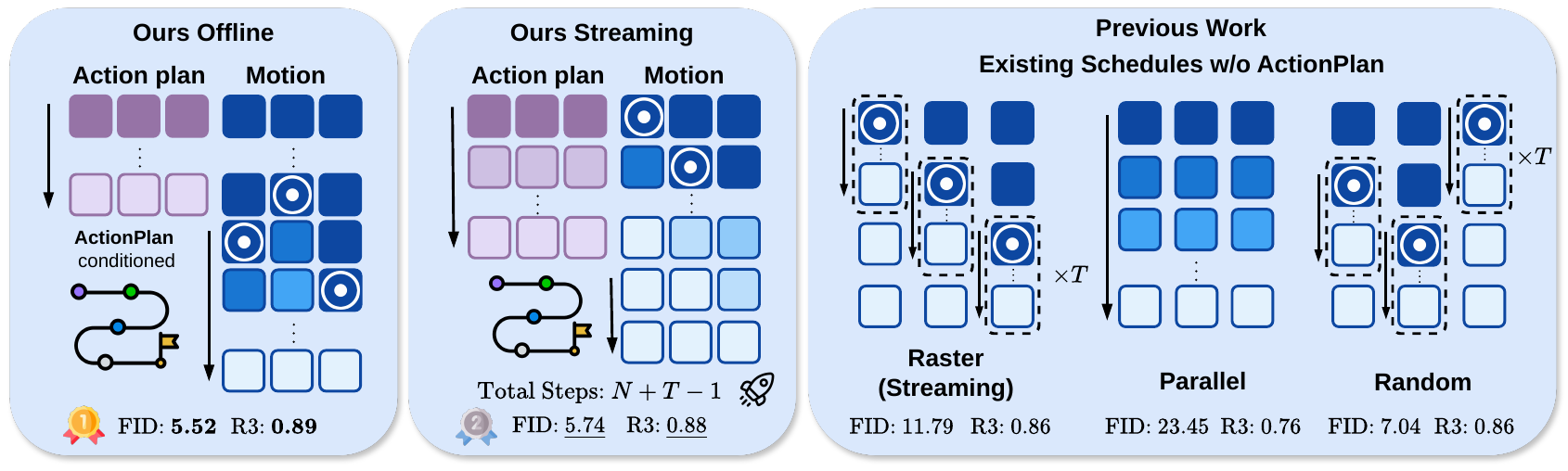}
\caption{\textbf{Comparison of generation paradigms}, where darker shading 
indicates higher noise levels. By introducing frame-level action plans 
as semantic conditioning, \methodName{} achieves significantly better 
FID and R-Precision compared to schedules without ActionPlan. 
Additionally, our streaming mode completes generation in only $N+T-1$ 
total steps ($N$: motion tokens, $T$: flow matching steps), enabling 
efficient low-latency generation without sacrificing motion quality.}
  \label{fig:streaming_compare}
  \vspace{-0.4cm}
\end{figure}

Existing methods largely fall into two specialized and mutually exclusive paradigms. 
Offline frameworks~\cite{barquero2024seamless,tevet2022human, chen2023executing} achieve high-quality results by leveraging global bidirectional attention to condition on the future with parallel or random order generation (c.f. Fig.~\ref{fig:streaming_compare}). 
However, their non-causal design requires access to the entire sequence, making them unsuitable for real-time streaming. 
In contrast, recent streaming-compatible models~\cite{xiao2025motionstreamer} achieve low-latency generation by adopting causal latent spaces and unidirectional raster order inference.
This strict causality, however, imposes a form of temporal myopia: without access to future context, streaming models often miss semantics from complex prompts, leading to less consistent motion generation. Furthermore, they are inherently incapable of bidirectional tasks such as editing and in-betweening.

In this work, we aim to preserve the low-latency benefits of streaming motion generation while equipping the model with awareness of global and future motion. We argue that effective online motion synthesis requires anticipating upcoming actions rather than reacting solely to past context.

To this end, we propose \textbf{\methodName} (c.f. Fig.~\ref{fig:teaser_figure}), a hierarchical framework that first predicts an action plan, a sequence of fine-grained, frame-level textual action descriptors, which subsequently conditions motion generation. By \emph{decoupling high-level semantic planning from low-level kinematic synthesis}, our approach enables state-of-the art generation quality and \emph{future-aware streaming without sacrificing accuracy}.
w
Action plans are temporally aligned with motion at each frame and their generation is trained jointly with motion latent generation using latent-specific diffusion timesteps for text and individual motion frames~\cite{chen2024diffusion, wewer25srm}. These independent timesteps enable flexible denoising schedules at inference time. We first generate action plans and then denoise the motion using task-specific timestep schedules tailored to online streaming, motion editing, or in-betweening.  As illustrated in~\cref{fig:streaming_compare}, this flexible scheduling, guided by action plans, allows offline and online generation by choosing different overlapping denoising orders.
We evaluate \methodName{} on HumanML3D-272~\cite{Guo2022CVPR_humanml3d, xiao2025motionstreamer}. 
In offline mode, \methodName{} improves on the best offline competing 
method~\cite{meng2025rethinking} by 22\% in FID. In online streaming 
mode, \methodName{} surpasses even the best offline 
baseline~\cite{meng2025rethinking} by 18\% and the best streaming 
method~\cite{xiao2025motionstreamer} by 51\% in FID, while achieving 
up to $5.25\times$ faster token generation.

We also conduct comprehensive ablations which validate that our action plan generation consistently improves performance in both setups and that our single model supports editing and in-betweening. Our code and pretrained model will be made publicly available. 

Our contributions are summarized as follows:
\begin{itemize}

\item We introduce \methodName, a hierarchical diffusion framework that decouples high-level action planning from low-level motion generation, by introducing varying noise levels for individual text and motion frames.

\item We propose to generate frame-level textual action plans first to 
strengthen the semantic conditioning for subsequent motion generation. Comprehensive ablations validate the effectiveness of our action plan.

\item We demonstrate that our action plan–guided sampling strategy supports both, offline and streaming generation, editing, and in-betweening within the same model without fine-tuning. Experiments show that our method significantly outperforms previous methods in both motion generation quality and inference speed. 
\end{itemize}
\section{Related Work}

\noindent\textbf{Text-to-motion generation.}
Text-conditioned motion generation aims to synthesize 3D human motions \emph{conditioned on} natural language descriptions~\cite{Guo2022CVPR_humanml3d}.
Early works learn cross-modal mappings by aligning text and motion in latent spaces~\cite{petrovich2023tmr, tevet2022motionclip, petrovich2022temos, ahuja2019language2pose}, or by adopting diffusion-based models~\cite{zhang2024motiondiffuse, tevet2022human, shafir2023human, yuan2023physdiff, kim2023flame, chen2023executing, li2025unimotion, dabral2022mofusion, zhang2023remodiffuse, zhang2023finemogen, karunratanakul2023gmd}
Following works~\cite{zhou2024emdm, zhang2024motion, huang2024stablemofusion, azadi2023make, zhao2024dartcontrol, zhang2024tedi, tevet2024closd, shi2024interactive, chen2024taming, zhang2024large, chen2025free, petrovich2024multi, liang2024intergen, zhangflashmo} mainly \emph{focus} on improving architectures or training strategies.
Another line of motion generation methods utilizes Vector Quantized Variational Autoencoders (VQ-VAE) to discretize motions into tokens~\cite{guo2022tm2t, zhang2023generating, jiang2023motiongpt, lu2023humantomato, lu2025scamo, zhou2024avatargpt, fan2025go, wan2024tlcontrol, wang2025motiondreamer, EgoLM, chen2025language, guo2025snapmogen, pinyoanuntapong2025maskcontrol, ghosh2025duetgen, zhang2025kinmo, liu2025gesturelsm},
these methods mainly apply BERT-style bidirectional transformers~\cite{fan2024textual, pinyoanuntapong2024mmm, guo2024momask, pinyoanuntapong2024bamm, cho2024discord} to predict the codebook entries via cross-entropy supervision, treating motions as a foreign language. 
More recent works~\cite{meng2025rethinking, meng2025absolute, xiao2025motionstreamer, zhu2025motiongpt3, tuautoregressive, he2026molingo} utilize continuous-valued motion latents to generate higher-quality human motions, these approaches still employ bidirectional attention mechanisms to denoise motion latents under textual guidance.
While such offline designs enable high-quality generation and support flexible tasks such as motion editing and in-betweening~\cite{pinyoanuntapong2024mmm, tevet2022human, dai2024motionlcm, pinyoanuntapong2025maskcontrol, kim2023flame, guo2024momask, meng2025rethinking}, they fundamentally rely on non-causal inference and assume access to the complete motion sequence.
Consequently, these methods are inherently unsuitable for low-latency interactive applications that require incremental, streaming capability.

\vspace{0.1cm}
\noindent\textbf{Real-time and online motion synthesis.}
Interactive applications require motion to be generated incrementally under streaming text input~\cite{shi2024interactive}.
Recent works apply diffusion models in an auto-regressive manner for real-time interactive character control, such as CAMDM~\cite{chen2024taming} and A-MDM~\cite{shi2024interactive}.
Ready-to-React~\cite{cen2025ready_to_react} further extends this idea to online two-character interaction.
In addition, CLoSD~\cite{tevet2024closd} and DART~\cite{zhao2024dartcontrol} explore real-time text-driven motion control with streaming prompts.
However, these approaches are often not strictly causal because they rely on a fixed-length context window. As generation progresses over time, access to earlier history diminishes and semantic coherence gradually fades.
More recently, MotionStreamer~\cite{xiao2025motionstreamer} proposes streaming text-to-motion generation with a continuous \emph{causal} latent space and auto-regressive denoising, enabling online response with variable-length historical context.
In contrast, we study a streaming-compatible formulation that additionally supports future-conditioned, bidirectional operations (e.g., editing and in-betweening) within a unified framework with action plans that clearly improve generation quality.

\vspace{0.1cm}
\noindent\textbf{Structured generation with diffusion.} 
In image generation, early representative methods~\cite{rombach2021highresolution} perform diffusion in a learned latent space. Recent approaches encode an image into a sequence of latents, enabling structured generation and explicitly exposing token ordering or masking as a design choice~\cite{li2024autoregressive, fan2024fluid}. Following this line, motion diffusion models adopt learned motion latents with masked auto-regressive generation to improve realism and text–motion alignment~\cite{meng2025rethinking, he2026molingo}.
For signals with strong spatial dependencies, SRMs~\cite{wewer25srm} predict an uncertainty-driven adaptive generation order, yielding consistent gains in both fidelity and correctness. More recently, Latent Forcing~\cite{baade2026latent} couples latent denoising with pixel-level denoising via separate noise schedules, producing an informative early ``blueprint'' and implicitly learning a underlying generation order for synthesis. A similar principle appears in motion generation: AutoKeyframe~\cite{zheng2025autokeyframe} generates keyframes first and then fills in the remaining frames.
Diffusion Forcing~\cite{chen2024diffusion} further generalizes structured sampling by assigning frame-specific noise levels, which facilitates flexible causal generation. Subsequent works leverage frame- or latent-wise timestep schedules together with tailored sampling strategies to enhance quality~\cite{cai2025flooddiffusion, yu2026causal}.
Motivated by these advances, we are the first method to generate textual action plans as a guiding blueprint and then perform structured real-time motion generation,
achieving higher speed and quality while remaining plug-and-play for downstream applications.
\section{\methodName{}: A Framework for Diverse Motion Tasks}

\begin{figure}[t]
  \centering
  \includegraphics[width=\textwidth]{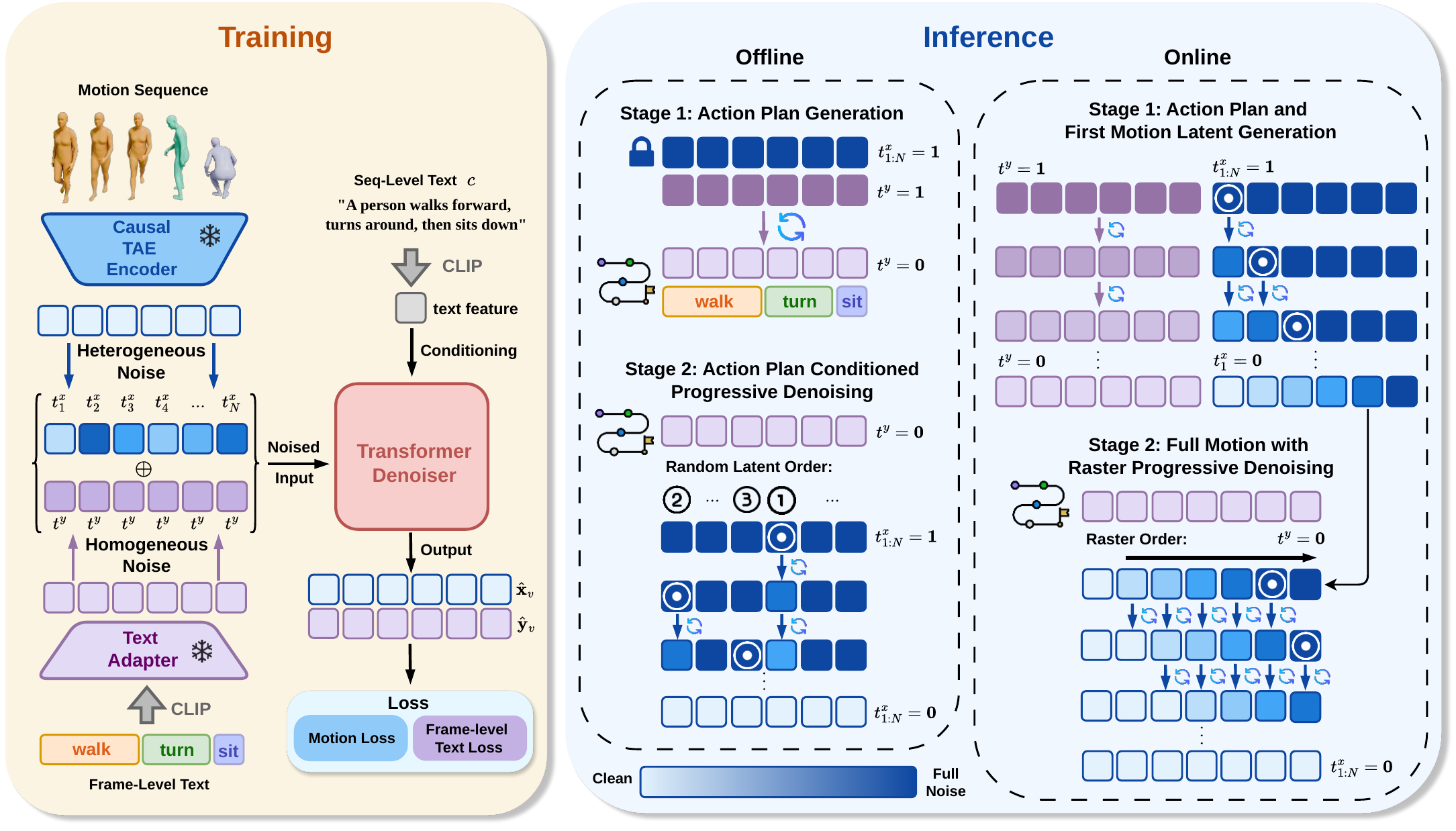}
  \caption{\textbf{Overview of our \methodName{}}. (a) During training, motion latents are noised with per-frame heterogeneous timesteps while frame-level text latents share a single global timestep. A Transformer Denoiser is trained to jointly reconstruct both. During inference, the model operates in two modes: in offline mode (b), 
the action plan is fully generated first and then motion latents are 
denoised in random pyramid order; in streaming mode (c), the action plan 
is denoised alongside the first motion frame, followed by raster 
progressive denoising of the remaining latents.
  }
  \label{fig:model}
  \vspace{-0.4cm}
\end{figure}

We propose \methodName{}, a unified text-to-motion (T2M) generation framework based on latent diffusion that supports versatile tasks via latent-specific diffusion timesteps. An overview of the proposed framework is shown in~\cref{fig:model}. Our key idea is to predict textual action plans that are aligned with motion at each frame and to allow independent denoising of each motion latent. At inference time, we first generate the per-frame action plan, then denoise the motion latents using task-specific schedules tailored to different downstream applications.
We first introduce the problem setup and motion autoencoder in~\cref{subsec:preliminary} and then describle co-generation of motion and text in~\cref{subsec:text-align}. We further discuss our latent-specific noise scheduling in~\cref{subsec:training} which allows flexible sampling guided by action plan generation in~\cref{subsec:sampling}. 

\subsection{Preliminaries}\label{subsec:preliminary}

\vspace{0.1cm} \noindent\textbf{Pose representation.}
We represent each pose
$m \in \mathbb{R}^{272}$ using SMPL-based 6D
rotations~\cite{zhou2019continuity, xiao2025motionstreamer} as:
\begin{equation}
  m = \{\dot{r}^x, \dot{r}^z, \dot{r}^a, j^p, j^v, j^r\},
\end{equation}
where $(\dot{r}^x, \dot{r}^z) \in \mathbb{R}^2$ are root linear velocities
on the ground plane, $\dot{r}^a \in \mathbb{R}^6$ is the root angular
velocity in 6D rotation, $j^p \in \mathbb{R}^{3K}$,
$j^v \in \mathbb{R}^{3K}$, and $j^r \in \mathbb{R}^{6K}$ are local joint
positions, velocities, and rotations respectively, with $K = 22$ joints. 
Unlike the widely used 263D representation~\cite{Guo2022CVPR_humanml3d}, which requires inverse kinematics to recover SMPL~\cite{SMPL:2015} parameters, the 272D representation directly drives the SMPL model without post-processing, avoiding conversion-induced rotation errors. We use this representation in all experiments.

\vspace{0.1cm} \noindent\textbf{Motion autoencoder.}
We adopt the Causal Temporal AutoEncoder (Causal TAE)~\cite{xiao2025motionstreamer} to encode raw motion sequences into a
continuous latent space. Given a motion sequence
$M = \{m_1, \dots, m_N\}$ with $m_t \in \mathbb{R}^{D}$ ($D = 272$),
the encoder $\mathcal{E}$ produces a sequence of latents
$\mathbf{x} = \{x_1, \dots, x_{N/l}\}$ with $x_i \in \mathbb{R}^{d_c}$,
where $l$ is the temporal downsampling rate and $d_c$ is the latent dimension.
Both the encoder and decoder are built from 1D causal convolutions, ensuring
that each latent depends only on current and past frames.
This causal
property enables online decoding: motion frames can be reconstructed
sequentially as latents become available, without waiting for the full
sequence. The Causal TAE is trained using standard VAE loss with an additional root stability loss, for more details we refer to the original work~\cite{xiao2025motionstreamer}. We keep the the TAE frozen for later training.

\subsection{Joint Diffusion on Action Plans and Kinematic Motion}\label{subsec:text-align}

We generate motion from action plans using diffusion in the Causal TAE latent space. Prior methods rely on a single sequence-level text description, offering only coarse conditioning and failing to capture the temporal structure of multi-action sequences (e.g., “walk forward then sit down”). Since both global intent and per-frame actions matter, we advocate for a hierarchical text representation by additionally predicting action plans aligned with motion at every frame. Thus, conditioned on a
sequence-level text description $c$ encoded by CLIP, our diffusion model $G_\theta$ learns to generate a sequence of action plans $\{y_1, ..., y_N\}$ with temporally aligned motion latent codes $\{x_1, ..., x_N\}$.

\vspace{0.1cm} \noindent\textbf{Action and motion representations.}  Action plans $\{y_1, ..., y_N\}$ are obtained by encoding per-frame descriptions with CLIP and compressing them via a pretrained MLP autoencoder to a 16-D space matching the motion latent dimension. Unlike the Causal TAE introduced in \cref{subsec:preliminary}, the action autoencoder operates independently per latent, preserving fine-grained temporal alignment. Training combines three objectives: an MSE reconstruction loss in the original embedding space, a neighbor-preservation loss that encourages correct action label retrieval, and a variance regularization term that discourages dimensional collapse by pushing each latent dimension toward unit variance. More details are provided in supplementary. 
Inspired by UniMotion~\cite{li2025unimotion}, the text latents are downsampled by a factor of 4 in the temporal axis and we align frame-level text with motion by concatenating the latents at each frame as one vector $(x_i, y_i)$.

\vspace{0.1cm} \noindent\textbf{Text and motion generation.}
Our denoiser $G_\theta$, a Transformer network, takes a sequence of concatenated vectors $\{(x_i, y_i)\}_1^N$ together with diffusion timesteps as input and predicts the velocity of the denoising trajectory: $\hat{\mathbf{x}}_v, \hat{\mathbf{y}}_v$. We train our diffusion network with carefully designed noise scheduling to allow flexible sampling at test time, which is introduced in the following section. 

\subsection{Training with Latent-specific Noise Levels}\label{subsec:training}
We assign each latent an independent noise level, allowing the model to handle heterogeneous denoising states, enabling conditioning on partially clean latents at inference and allowing flexible sampling strategies.

\vspace{0.1cm} \noindent\textbf{Heterogeneous noise scheduling.} We assign each motion latent $x_i$ an independent timestep $t_i^x\in [0,1]$, sampled according to the $\bar{t}$ algorithm~\cite{wewer25srm}, avoiding collapse of the mean to a Bates distribution (c.f. appendix for details). For action plan latents $y_i$, a single global timestep $t^y \sim \mathcal{U}(0,1)$ is sufficient and reduces complexity. 
We follow Rectified Flow~\cite{liu2023flow} with velocity parameterization, i.e. our learned denoising function $G_\theta$ is formulated as:
\begin{equation}
  (\hat{\mathbf{x}}_v, \hat{\mathbf{y}}_v)
  = G_\theta(\mathbf{x}_{\mathbf{t}^x}, \mathbf{y}_{t^y};\,
    \mathbf{t}^x, t^y, c),
    \label{eq:denoiser}
\end{equation}
where $\mathbf{t}^x = \{t_1^x, \dots, t_N^x\}$ and $\mathbf{x}_{\mathbf{t}^x}, \mathbf{y}_{t^y}$ are noisy states of motion and action latents, according to the Rectified Flow forward process~\cite{liu2023flow}.

\vspace{0.1cm} \noindent\textbf{Training with mixed datasets.} To supervise action plans $y_i$, we require a dataset with both sequence level text $c$ and frame-level text annotations. HumanML3D-272~\cite{xiao2025motionstreamer} provides sequence level text while BABEL-272~\cite{BABEL:CVPR:2021, xiao2025motionstreamer} additionally provides frame-level annotations for a subset of samples. To leverage the best of two datasets, we propose mixed training using both datasets with masking. The standard Rectified Flow loss for text prediction is $\mathcal{L}_\text{text}=\left\| \hat{\mathbf{y}}_v - (\mathbf{y}_0 - \mathbf{\epsilon})\right\|^2_2$, where $\mathbf{\epsilon} \sim \mathcal{N}(0, I)$ is the pure noise and $\mathbf{y}_0$ the clean latents. We introduce an indicator $w \in \{0, 1\}$ to dynamically disable the text loss: 
\begin{equation}
    \mathcal{L}_\text{text} =w\left\|  \hat{\mathbf{y}}_v - (\mathbf{y}_0 - \mathbf{\epsilon}) \right\|^2_2
    \label{eq:masked_loss}
\end{equation}
where $w = 1$ if frame-level text annotations are available for this sequence. 
The training objective of our model is then the reconstruction loss for motion and text, and per-latent variance loss $\mathcal{L}_\text{var}$~\cite{wewer25srm}: 
\begin{equation}
  \mathcal{L} = \lambda_x \, \mathcal{L}_{\text{motion}}
              + \lambda_y \, \mathcal{L}_{\text{text}},
    \label{eq:loss-full}
\end{equation}
where $\mathcal{L}_{\text{motion}} = \left\| \hat{\mathbf{x}}_v - (\mathbf{x}_0 - \mathbf{\epsilon}) \right\|^2_2$ and $\mathbf{x}_0$ denotes the clean motion latents.

\subsection{Flexible Sampling with Action Plan Generation}\label{subsec:sampling}

The flexible denoising formulation in~\cref{subsec:training} enables inference-time sampling strategies with independently scheduled text and motion timesteps at each frame. Since global semantic structure, \ie, what action occurs when, is crucial for maintaining consistency in long and complex motions, we first generate an action plan and then synthesize the full motion using overlapping denoising windows with random (offline) or raster (streaming) orders.

\vspace{0.1cm} \noindent\textbf{Stage 1: Action plan generation}. We keep all motion latents at pure noise ($t_{i}^{x} = 1, \forall i$) and denoise the action latents $\mathbf{y}$ over $T$ steps: $t^{y}_{s} = 1 - s\,\Delta t, \quad s = 0, \dots, T$. At the end of this stage we obtain clean frame-level text latents $\hat{\mathbf{y}}_{0}$, the \textit{action plan}, which specify what action occurs at each temporal position before any motion is generated. 

\vspace{0.1cm} \noindent\textbf{Stage 2: Progressive denoising schedule.}
Given the action plan, we generate the full motion while conditioning on frame-level semantics. 
Although sequentially denoising frame-after-frame is a natural solution, it is computationally prohibitive, see~\cref{fig:streaming_compare}. 
Instead, we overlap the denoising intervals of individual frames~\cite{chen2024diffusion, wewer25srm}: motion frames are partitioned into a denoising set and a waiting set. After a few iterations, one additional noisy latent is moved from waiting set to the denoising set (in ~\cref{fig:streaming_compare} and~\cref{fig:model} such activation is indicated as a white dot). Thus, multiple active latents are updated simultaneously with individual denoising progress, increasing throughput.

\textbf{Offline:} Motion latents are activated in random order without temporal constraints. We find that random activation yields the best generation quality.

\textbf{Online / Streaming:} Latents are activated sequentially in temporal order. While such causal activation typically leads to degraded motion quality due to limited future context, our action plan conditioning preserves global semantic consistency during streaming and heavily reduces this degradation. In online mode, we denoise the action plan alongside the first motion latent.

\section{Experiments}
In this section, we compare \methodName{} with state-of-the-art (SOTA) T2M generation methods, ablate key design choices and showcase downstream applications enabled by our flexible architecture. The results show that \methodName{} outperforms all prior methods in both online streaming and offline mode, and that our action plan generation improves motion quality.

\subsection{Experimental Setup}

\begin{figure}[th]
  \centering
  \includegraphics[width=1.0\linewidth]{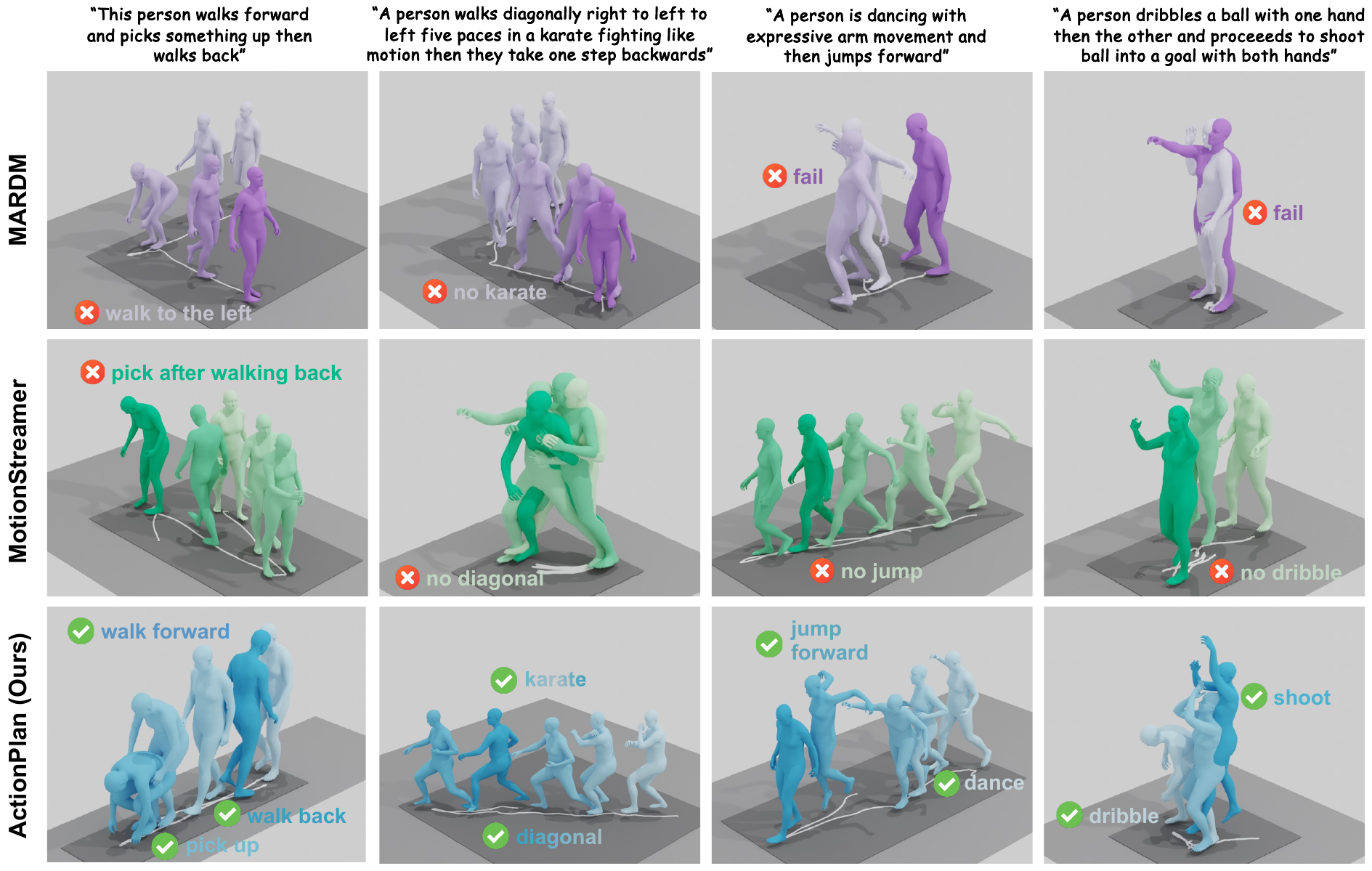}
  \caption{\textbf{Qualitative comparison} with MARDM~\cite{meng2025rethinking} 
and MotionStreamer~\cite{xiao2025motionstreamer} on four text prompts. The color varies from light to dark representing the time flow.
Incorrectly generated or missing actions are marked with {\color{red}$\times$}, 
and correctly executed actions with {\color{green}$\checkmark$}. 
By generating a frame-level action plan prior to motion synthesis, 
\methodName{} faithfully executes all specified actions in the correct 
order, while baselines frequently miss or misorder key actions. See Supp. for videos.}
  \label{fig:qualitative}
  \vspace{-0.5cm}
\end{figure}

\vspace{0.1cm} \noindent\textbf{Dataset.}
We follow the evaluation protocol introduced by MotionStreamer~\cite{xiao2025motionstreamer} and conduct experiments on HumanML3D-272 \cite{xiao2025motionstreamer} 
with additional frame-level text labels from BABEL~\cite{BABEL:CVPR:2021}. MotionStreamer corrects inaccuracies in original HumanML3D~\cite{Guo2022CVPR_humanml3d} by using 272-D pose representation and resamples all motions to 30 FPS, yielding higher-quality data. BABEL includes fine-grained frame-level text annotation and partially overlaps with the motion sequences in HumanML3D. We train our method with the same set of training motions from HumanML3D-272 as used in previous methods~\cite{xiao2025motionstreamer, Guo2022CVPR_humanml3d} and query frame-text labels when they are available from BABEL to supervise action plan generation. 

\vspace{0.1cm} \noindent\textbf{Evaluation metrics.}
We report a comprehensive set of metrics following~\cite{Guo2022CVPR_humanml3d, xiao2025motionstreamer}.
To assess generation quality, we measure Fréchet Inception Distance
(FID)~\cite{he2016deep}.
For semantic alignment, we report R-Precisions and
Multi-modal Matching Score (MatchS), which evaluate retrieval accuracy and feature-space proximity between generated motions and their text descriptions, respectively~\cite{Guo2022CVPR_humanml3d}. 
We additionally measure Diversity~\cite{Guo2022CVPR_humanml3d} to quantify variation across generated samples. 

\vspace{0.1cm} \noindent\textbf{Implementation details.}
The latent denoising model is implemented as a lightweight Transformer~\cite{vaswani2017attention} with 16 layers, 16 attention heads, and a hidden dimension of 1024 . We use CLIP to embed both frame level text and sequence-level text. For the frame level text, we use an autoencoder to reduce the original textual dimension to 16. In order to retrieve the text from the latent space, we decode the 16-d latents and use a simple KNN method on the CLIP embeddings, following~\cite{li2025unimotion}. 
During inference, we use 2 denoising steps for progressive sampling (\cref{tab:ablation_sampling_strategy}) and 25 total denoising steps for rectified flow.

\subsection{Baseline Comparison on Text-to-motion Generation}\label{subsec:baseline}

\vspace{0.1cm} \noindent\textbf{Quantitative evaluation.}
We follow the training and evaluation setup of
<MotionStreamer~\cite{xiao2025motionstreamer}: all models are trained on the HumanML3D-272~\cite{Guo2022CVPR_humanml3d, xiao2025motionstreamer} training set, and evaluated with their pretrained TMR-based evaluator~\cite{xiao2025motionstreamer, petrovich2023tmr}. All baselines are re-trained from scratch using their own official implementations and the same 272-D representation for fair comparison.

As shown in~\cref{tab:t2m}, a single \methodName{} checkpoint supports both offline and 
streaming generation simply by switching the sampling strategy at inference time, no 
retraining or fine-tuning is required. Both variants consistently outperform all competing 
methods across all metrics, including diffusion-~\cite{tevet2022human, chen2023executing}, VQ-~\cite{zhang2023generating, jiang2023motiongpt, guo2024momask}, streaming-~\cite{xiao2025motionstreamer}, 
and masked auto-regressive~\cite{meng2025rethinking} approaches. Notably, our offline variant achieves substantial gains over the best baseline, with 22\% improvement in FID and R-Precision@3 reaching 0.892. Our online mode, being comparable to our offline mode, is 51\% better in FID than the best streaming method MotionStreamer~\cite{xiao2025motionstreamer}. 

\begin{table}[t]
\centering
\footnotesize 
\setlength{\tabcolsep}{4.5pt} 
\renewcommand{\arraystretch}{1.1}
\begin{tabular}{@{} l c c c c c c @{}}
\toprule
\multirow{2}{*}{\textbf{Method}} & \multirow{2}{*}{\textbf{FID} $\downarrow$} & \multicolumn{3}{c}{\textbf{R-Precision}\,$\uparrow$} & \multirow{2}{*}{\textbf{MatchS} $\downarrow$} & \multirow{2}{*}{\textbf{Div} $\rightarrow$} \\
\cline{3-5}
& & \textbf{Top 1} & \textbf{Top 2} & \textbf{Top 3} & & \\ 
\midrule
Real & 0.002 & 0.702 & 0.864 & 0.914 & 15.151 & 27.492 \\
\midrule
MDM~\cite{tevet2022human} & 23.454 & 0.523 & 0.692 & 0.764 & 17.423 & 26.325 \\
MLD~\cite{chen2023executing} & 18.236 & 0.546 & 0.730 & 0.792 & 16.638 & 26.352 \\
T2M-GPT~\cite{zhang2023generating} & 12.475 & 0.606 & 0.774 & 0.838 & 16.812 & 27.275 \\
MotionGPT~\cite{jiang2023motiongpt} & 14.375 & 0.456 & 0.598 & 0.628 & 17.892 & 27.114 \\
MoMask~\cite{guo2024momask} & 12.232 & 0.621 & 0.784 & 0.846 & 16.138 & 27.127 \\
AttT2M~\cite{zhong2023attt2m} & 15.428 & 0.592 & 0.765 & 0.834 & 15.726 & 26.674 \\
MotionStreamer~\cite{xiao2025motionstreamer} & 11.790 & 0.631 & 0.802 & 0.859 & 16.081 & {27.284} \\
MARDM~\cite{meng2025rethinking} & {7.044} & {0.669} & 0.806 & 0.860 & 15.892 & 26.235 \\
\midrule
\textbf{\methodName{}-Streaming} & \underline{5.735} & \underline{0.672} & \underline{0.822} & \underline{0.877} & \underline{15.315} & \textbf{27.287} \\
\textbf{\methodName{}-Offline} & \textbf{5.522} & \textbf{0.687} & \textbf{0.838} & \textbf{0.892} & \textbf{15.09} & \underline{27.272} \\
\bottomrule

\end{tabular}

\vspace{2mm} 
\caption{\textbf{Quantitative comparison} with SOTA T2M generation methods on HumanML3D-272~\cite{Guo2022CVPR_humanml3d, xiao2025motionstreamer} test set. MatchS and Div denote the matching score and diversity respectively. \textbf{Bold} indicates best results, \underline{underline} indicates second best. \methodName{} achieves better performance in both offline and online-streaming mode, maintaining both efficiency and high quality motion generation.
}
\label{tab:t2m}
  \vspace{-0.6cm}
\end{table}

\vspace{0.1cm} \noindent\textbf{Runtime performance.}
To evaluate practical efficiency, we measure the latency for generating one motion token (4 frames) and compare our method in online streaming mode with baseline MARDM~\cite{meng2025rethinking} and MotionStreamer~\cite{xiao2025motionstreamer} in the table below (unit: ms). All experiments are done on a single NVIDIA A100 GPU. 
\begin{wraptable}{r}{0.42\linewidth}
\vspace{-12pt}
\centering
\small
\setlength{\tabcolsep}{2pt}
\begin{tabular}{lcc}
\toprule
Method & First $\downarrow$ & Others $\downarrow$ \\
\midrule
MARDM & 210 & 210 \\
MotionStreamer & 360 & 360 \\
\midrule
Ours & \textbf{146} & \textbf{40} \\
\bottomrule
\end{tabular}
\vspace{-12pt}
\end{wraptable}
MARDM~\cite{meng2025rethinking} and MotionStreamer~\cite{xiao2025motionstreamer} generate motions in an autoregressive manner which is more efficient than full motion sampling yet they still require full denoising for every token. In contrast, our overlapping schedule requires a small overhead for the first token but is significantly faster on the following tokens, achieving a 1.44$\times$--2.47$\times$ speedup for the initial token and 5.25$\times$--9$\times$ during continuous streaming. Notably, our streaming mode achieves significant speedup compared to baselines while maintaining much better motion quality, demonstrating robust adaptability of our method.

\vspace{0.1cm}\noindent\textbf{Qualitative comparisons.} \cref{fig:qualitative} 
presents qualitative comparisons with MotionStreamer~\cite{xiao2025motionstreamer} 
and MARDM~\cite{meng2025rethinking} on the text-to-motion task.
Despite its expensive masked auto-regressive inference, MARDM frequently 
fails to capture fine-grained semantic details: given the prompt 
\textit{``a person walks forward and picks something up then walks 
back''}, it generates a leftward walk without the pick-up action. 
MotionStreamer, as a strictly causal streaming method, suffers from 
action drift and incompleteness due to the lack of future context: 
it omits the dribbling action in \textit{``a person dribbles a ball 
then shoots into a goal''} and misorders actions in sequential prompts. 
In contrast, \methodName{} faithfully executes all specified actions 
in the correct order across all examples, demonstrating that the 
frame-level action plan effectively grounds generation to the full 
semantic content of the input prompt.

\subsection{Ablation Studies}
We conduct comprehensive ablation studies to validate the key design choices of our framework and report results in \cref{tab:ablation_training_strategy} and \cref{tab:ablation_sampling_strategy}. All methods are evaluated on the same HumanML3D-272~\cite{xiao2025motionstreamer} test set. 

\begin{table}[t]
\centering
\renewcommand{\arraystretch}{1.3} 
\setlength{\tabcolsep}{5pt}
\resizebox{\textwidth}{!}{%
\begin{tabular}{@{} l | c | c | c | c c c c c c @{}}
\toprule
\multirow{2}{*}{\textbf{Configurations}} & \textbf{Training} & \textbf{ActionPlan} & \multirow{2}{*}{\textbf{Mode}} & \multirow{2}{*}{\textbf{FID} $\downarrow$} & \multicolumn{3}{c}{\textbf{R-Precision}\,$\uparrow$} & \multirow{2}{*}{\textbf{MatchS}\,$\downarrow$} & \multirow{2}{*}{\textbf{Div} $\rightarrow$} \\
\cline{6-8}
 & \textbf{data} & \textbf{generation} & & & \textbf{Top 1} & \textbf{Top 2} & \textbf{Top 3} & & \\
\hline
Real motion & & &  & 0.002 & 0.702 & 0.864 & 0.914 & 15.151 & 27.492 \\
\hline
\multirow{2}{*}{\textbf{(A)} No frame-level text} 
    & \multirow{2}{*}{partial} & \multirow{2}{*}{\ding{55}} 
    & Offline   & 9.674  & 0.634 & 0.790 & 0.851 & 15.764 & 27.207 \\
 &  &  & Streaming & 10.693 & 0.627 & 0.785 & 0.848 & 15.931 & 27.212 \\
\hline
\multirow{2}{*}{\textbf{(B)} Frame-level text w/ actionplan} 
    & \multirow{2}{*}{partial} & \multirow{2}{*}{\checkmark} 
    & Offline   & 7.863 & 0.625 & 0.791 & 0.854 & 15.849 & 27.232 \\
 &  &  & Streaming & 8.018 & 0.625 & 0.789 & 0.854 & 15.866 & 27.220 \\
\hline
\multirow{2}{*}{\textbf{(C)} No frame-level text} 
    & \multirow{2}{*}{full} & \multirow{2}{*}{\ding{55}} 
    & Offline   & 6.952 & 0.661 & 0.814 & 0.873 & 15.390 & \underline{27.275} \\
 &  &  & Streaming & 8.648 & 0.655 & 0.810 & 0.869 & 15.586 & \textbf{27.287} \\
\hline
\textbf{(D)} Frame-level text T\&M co-gen 
    & full & \ding{55} 
    & Offline   & 6.093 & 0.677 & 0.822 & 0.878 & 15.234 & 27.186 \\
\hline
\multirow{2}{*}{\textbf{(E)} Frame-level text w/ actionplan} 
    & \multirow{2}{*}{full} & \multirow{2}{*}{\checkmark} 
    & Offline   & \textbf{5.522} & \textbf{0.687} & \textbf{0.838} & \textbf{0.892} & \textbf{15.086} & 27.272 \\
 &  &  & Streaming & \underline{5.878} & \underline{0.675} & \underline{0.821} & \underline{0.875} & \underline{15.369} & 27.228 \\
\bottomrule
\end{tabular}}
\vspace{2mm}
\caption{\textbf{Ablation studies} for both Offline and Streaming modes. 
Training on the full dataset (C--E) consistently outperforms the partial intersection subset (A--B), validating our masked loss design. 
Frame-level text prediction (D, E) improves over no frame text (C). 
Action plan generation (E) further outperforms joint co-generation (D). 
These gains hold across both inference modes. \textbf{Bold}: best, \underline{underline}: second best.}
\label{tab:ablation_training_strategy}
  \vspace{-0.6cm}
\end{table}

\vspace{0.1cm} \noindent\textbf{Frame-level text semantics prediction.} We propose to additionally predict text for each frame, enabling more comprehensive semantic alignment between text and motion. We train models with and without frame text prediction on the full HumanML3D-272 training set. Even with simple joint generation of per-frame text and motion, our frame-level text prediction (\cref{tab:ablation_training_strategy}D) already improves over no text prediction (\cref{tab:ablation_sampling_strategy}C). 

\vspace{0.1cm} \noindent\textbf{Mixed dataset training.} To train on sequences that do not have frame-level text annotation, we propose a simple mask loss (~\cref{eq:masked_loss}) to fully leverage the complete HumanML3D~\cite{Guo2022CVPR_humanml3d} dataset. Compared to models (\cref{tab:ablation_training_strategy}A, B) trained on the overlapping subset between HumanML3D and BABEL, only 30\% of the full data, models trained on full data (\cref{tab:ablation_training_strategy}C, E) are consistently better. This underscores the importance of data scale and effective curation strategies.

\vspace{0.1cm} \noindent\textbf{Action Plan generation.} At inference time, we propose to first generate action plans and then complete the full motion. We show in \cref{tab:ablation_training_strategy} that this two-stage generation (\cref{tab:ablation_training_strategy}E) is better than simple joint generation (\cref{tab:ablation_training_strategy}D) of both text and motion with single timestep. 
This suggests that making motion aware of the future through action plans is crucial for good performance.

\vspace{0.1cm} \noindent\textbf{Progressive sampling strategy.} To support diverse tasks while maintaining efficiency, we propose to denoise in a pyramid fashion: gradually add one more latent to denoise multiple motion latents together, each with different timesteps. This independent timestep improves efficiency while maintaining flexibility. Alternatively, one can change the overlap window size $K=2$ between two consecutively added latents to enable faster (more latents are denoised at the same time) or slower (less latents are denoised at the same time). We ablate this design choice in \cref{tab:ablation_sampling_strategy}. It can be seen that the chosen size $K=2$ achieves the best balance between motion quality and text alignment. 

\begin{table}[t] 
\centering
\renewcommand{\arraystretch}{1.3} 
\setlength{\tabcolsep}{5pt}
\resizebox{\textwidth}{!}{%
\begin{tabular}{@{} l c c c c c c @{}}
\toprule
\multirow{2}{*}{\textbf{Progressive Sampling}} & \multirow{2}{*}{\textbf{FID}$\downarrow$} & \multicolumn{3}{c}{\textbf{R-Precision}$\uparrow$} & \multirow{2}{*}{\textbf{MatchS}$\downarrow$} & \multirow{2}{*}{\textbf{Div}$\rightarrow$} \\
\cline{3-5}
 & & \textbf{Top 1} & \textbf{Top 2} & \textbf{Top 3} \\
\midrule
Real & 0.002 & 0.702 & 0.864 & 0.914 & 15.151 & 27.492 \\
\midrule
Fully overlap (Parallel) & \textbf{5.420} & 0.681 & \underline{0.836} & \underline{0.886} & \underline{15.104} & 26.989 \\

2-step non-overlap & \underline{5.522} & \textbf{0.687} & \textbf{0.838} & \textbf{0.892} & \textbf{15.086} & \textbf{27.272} \\
5-steps non-overlap & 5.721 & \textbf{0.687} & 0.834 & 0.885 & 15.128 & 27.179 \\
10-steps non-overlap & 5.605 & 0.685 & 0.832 & 0.885 & 15.128 & \underline{27.261}\\
15-steps non-overlap & 5.612 & \underline{0.686} & 0.834 & 0.883 & 15.175 & 27.113 \\
Fully non-overlap (Random) & 5.566 & 0.683 & 0.828 & 0.880 & 15.160 & 27.138 \\

\bottomrule
\end{tabular}%
}
\vspace{2mm}
\caption{\textbf{Ablation on sampling strategy.} All rows use the same trained model with a fixed total of 25 denoising steps, differing only in the overlap between consecutively activated motion latents, ranging from fully parallel activation to fully sequential (non-overlapping) denoising. Our chosen schedule denoises each activated latent for 2 steps before activating the next, achieving the best trade-off between motion quality (FID) and text-motion alignment (R-Precision).
}
\label{tab:ablation_sampling_strategy}
  \vspace{-0.6cm}
\end{table}

\subsection{User Study}\label{subsec:user_study}
To further evaluate qualitative results, we conduct user studies on 
randomly selected HumanML3D-272 test prompts to evaluate 
text-to-motion generation and long motion generation. For each question, we present side-by-side animations in randomized order, asking them to select the best motion 
based on two criteria: (1)~\textit{semantic consistency}, how well the 
motion matches the given text description; and (2)~\textit{realism}, 
how natural, realistic, and smooth the motion appears.

\noindent\textbf{Text-to-motion generation.} We compare 
against MotionStreamer~\cite{xiao2025motionstreamer} and 
MARDM~\cite{meng2025rethinking} on $20$ prompts and release the user study to 30 participants. \methodName{} is 
preferred by $67.5$\% of participants, with MARDM and MotionStreamer 
receiving $12.2$\% and $20.3$\% of preferences respectively.

\noindent\textbf{Long motion generation.} We randomly select 20 long-horizon sequences, each composed of multiple prompts and compare against MotionStreamer~\cite{xiao2025motionstreamer} with 30 participants. \methodName{} is preferred by 
67.7\% of participants over 32.3\% for MotionStreamer.

\noindent \methodName{} is consistently preferred by participants by a large margin in both user studies, closely matching the quantitative improvements reported in \cref{tab:t2m}. The strong alignment between human judgments and automatic metrics highlights not only superior motion quality, realism, and smoothness, but also more accurate semantic alignment with the input text. Together, these results provide compelling evidence that our method substantially outperforms prior approaches across both perceptual and objective evaluations.

\begin{figure}[h]
  \centering
  \includegraphics[width=0.95\linewidth]{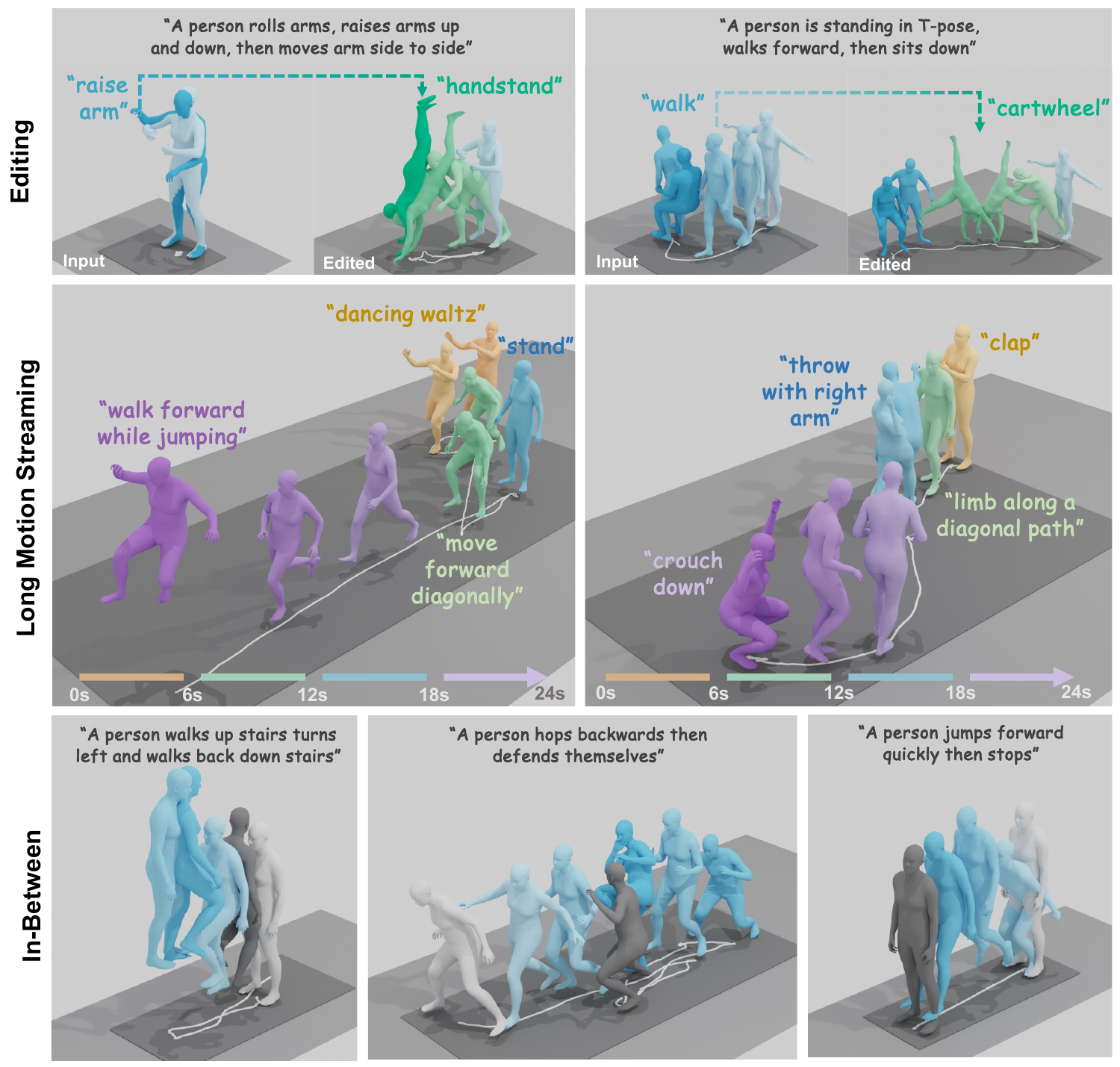}
  \caption{\textbf{\methodName{} supports diverse downstream applications zero-shot}. Darker shading indicates later time steps. 
\textbf{Editing} (top): regenerates selected latents conditioned on a new prompt ({\color{green}green}) while preserving others. 
\textbf{Long motion streaming} (middle): generates coherent long-horizon motion in successive chunks across prompts.
\textbf{In-betweening} (bottom): given fixed start (white) and end (dark grey) poses, fills in the intermediate motion. See Supp. for video results.}
  \label{fig:applications}
  \vspace{-0.5cm}
\end{figure}
\subsection{Applications}\label{subsec:applications}

Benefiting from our flexible training strategy, \methodName{} supports multiple generation modes without retraining or fine-tuning. We present three downstream applications below and refer readers to the supplementary for video results.

 \noindent\textbf{Motion editing.}
Given a motion sequence generated by \methodName{}, we fix a subset of latents and regenerate the rest conditioned on them and a new prompt. This is achieved by setting the fixed latent’s timesteps to $t_i^x = 0$ during sampling. As shown in \cref{fig:applications}, the edited regions reflect the new intent while smoothly blending with preserved segments.

\noindent\textbf{Motion in-betweening.}
Given fixed start and end poses and a text prompt, \methodName{} fills 
in the intermediate motion by treating boundary frames as fully denoised 
anchors and applying progressive denoising the remaining latents conditioned on both the boundaries and text. As shown in \cref{fig:applications}, the infilled motions are temporally smooth, and faithful to the input prompt.

\noindent\textbf{Long motion generation.} Given a sequence of prompts, \methodName{} generates long-horizon motion in successive chunks, each conditioned on its prompt and the previous chunk’s final tokens for continuity. As shown in \cref{fig:applications}, it produces temporally smooth motions that follow the intended action sequence.
\section{Conclusion and Limitations}

We presented \methodName{}, a hierarchical diffusion framework that 
unifies offline and streaming text-to-motion generation within a single 
model. By decoupling frame-level action planning from motion synthesis 
and introducing latent-specific noise schedules, \methodName{} enables 
future-aware semantic conditioning across a diverse set of tasks, such as streaming 
generation, offline generation, motion editing, and in-betweening, all in one single model without fine tuning. 

Experiments on HumanML3D-272 demonstrate that our offline mode outperforms prior methods by 21\% in FID and online mode achieves 51\% improvement in FID compared to the best streaming method while being $5.2\times$ faster. Our method is consistently favored by human evaluators, showing more than 67\% preference in all user studies. We further conduct comprehensive ablations which validate the importance of our action plan generation in various setups. Results also show that our method enables zero-shot applications for diverse tasks while keeping the generated motions smooth and faithful to diverse text control. Our code and model will be publicly released. 

\vspace{0.1cm} \noindent\textbf{Limitations.}
While \methodName{} enables flexible text-to-motion generation and 
supports multiple downstream tasks, there are several limitations. Firstly, it cannot yet model finger 
articulation or facial expressions which requires additional datasets with paired text descriptions for hands and faces. Secondly, our method generates human motion without scene or object awareness which limits its application in interactive agents or robotics that require models to actually interact with environment to finish complex daily tasks. We leave these for future works. 

\clearpage
{\small
\noindent\textbf{Acknowledgments:} Special thanks RVH members for the help and discussion. Prof. Gerard Pons-Moll is member of the Machine Learning Cluster of Excellence, EXC number 2064/1 - Project number 390727645. Gerard Pons-moll is endowed by the Carl Zeiss Foundation. Jan Eric Lenssen is supported by the German Research Foundation (DFG) - 556415750 (Emmy Noether Programme, project: Spatial Modeling and Reasoning). Xianghui Xie is partially funded by NVIDIA fellowship.

EN and CL contributed equally as joint first authors; they are allowed to change their order freely on their resume and website. CL initialized the core idea, organized the project, co-developed the method, co-supervised the experiments, and wrote the draft along with figures. EN co-developed the method, led the implementation of most prototypes and demo development, and conducted most experiments. YH co-supervised the experiments, contributed to draft refining and result visualization, including the teaser and result figures. XX co-supervised the experiments, contributed to draft refining and the application demo.
}

\bibliographystyle{splncs04}
\bibliography{main}

@String(CVPR  = {IEEE Conf. Comput. Vis. Pattern Recog.})

@String(NeurIPS = {Adv. Neural Inform. Process. Syst.})

@String(ICML  = {Int. Conf. Mach. Learn.})

@String(ICLR  = {Int. Conf. Learn. Represent.})

@String(AAAI  = {AAAI})

@String(TOG   = {ACM Trans. Graph.})

@String(CVPR  = {CVPR})

@String(NeurIPS = {NeurIPS})

@String(ICML  = {ICML})

@String(ICLR  = {ICLR})

@String(TOG   = {ACM TOG})

@inproceedings{guo2024momask,
  title={Momask: Generative masked modeling of 3d human motions},
  author={Guo, Chuan and Mu, Yuxuan and Javed, Muhammad Gohar and Wang, Sen and Cheng, Li},
  booktitle={Proceedings of the IEEE/CVF Conference on Computer Vision and Pattern Recognition},
  pages={1900--1910},
  year={2024}
}

@inproceedings{dai2024motionlcm,
  title={Motionlcm: Real-time controllable motion generation via latent consistency model},
  author={Dai, Wenxun and Chen, Ling-Hao and Wang, Jingbo and Liu, Jinpeng and Dai, Bo and Tang, Yansong},
  booktitle={European Conference on Computer Vision},
  pages={390--408},
  year={2024},
  organization={Springer}
}

@inproceedings{yu2026causal,
      title={Causal Motion Diffusion Models for Autoregressive Motion Generation},
      author={Yu, Qing and Watanabe, Akihisa and Fujiwara, Kent},
      booktitle={CVPR},
      year={2026}
  }

@inproceedings{BABEL:CVPR:2021,
  title = {{BABEL}: Bodies, Action and Behavior with English Labels},
  author = {Punnakkal, Abhinanda R. and Chandrasekaran, Arjun and Athanasiou, Nikos and Quiros-Ramirez, Alejandra and Black, Michael J.},
  booktitle = {Proceedings IEEE/CVF Conf.~on Computer Vision and Pattern Recognition (CVPR)},
  pages = {722--731},
  month = jun,
  year = {2021},
  doi = {},
  month_numeric = {6}
}

@inproceedings{meng2025rethinking,
  title={Rethinking Diffusion for Text-Driven Human Motion Generation: Redundant Representations, Evaluation, and Masked Autoregression},
  author={Meng, Zichong and Xie, Yiming and Peng, Xiaogang and Han, Zeyu and Jiang, Huaizu},
  booktitle={Proceedings of the Computer Vision and Pattern Recognition Conference},
  pages={27859--27871},
  year={2025}
}

@article{xiao2025motionstreamer,
  title={MotionStreamer: Streaming Motion Generation via Diffusion-based Autoregressive Model in Causal Latent Space},
  author={Xiao, Lixing and Lu, Shunlin and Pi, Huaijin and Fan, Ke and Pan, Liang and Zhou, Yueer and Feng, Ziyong and Zhou, Xiaowei and Peng, Sida and Wang, Jingbo},
  journal={arXiv preprint arXiv:2503.15451},
  year={2025}
}

@inproceedings{barquero2024seamless,
  title={Seamless human motion composition with blended positional encodings},
  author={Barquero, German and Escalera, Sergio and Palmero, Cristina},
  booktitle={Proceedings of the IEEE/CVF Conference on Computer Vision and Pattern Recognition},
  pages={457--469},
  year={2024}
}

@inproceedings{li2025unimotion,
  title={Unimotion: Unifying 3d human motion synthesis and understanding},
  author={Li, Chuqiao and Chibane, Julian and He, Yannan and Pearl, Naama and Geiger, Andreas and Pons-Moll, Gerard},
  booktitle={2025 International Conference on 3D Vision (3DV)},
  pages={240--249},
  year={2025},
  organization={IEEE}
}

@article{shafir2023human,
  title={Human motion diffusion as a generative prior},
  author={Shafir, Yonatan and Tevet, Guy and Kapon, Roy and Bermano, Amit H},
  journal={arXiv preprint arXiv:2303.01418},
  year={2023}
}

@article{tevet2022human,
  title={Human motion diffusion model},
  author={Tevet, Guy and Raab, Sigal and Gordon, Brian and Shafir, Yonatan and Cohen-Or, Daniel and Bermano, Amit H},
  journal={arXiv preprint arXiv:2209.14916},
  year={2022}
}

@article{meng2025absolute,
  title={Absolute Coordinates Make Motion Generation Easy},
  author={Meng, Zichong and Han, Zeyu and Peng, Xiaogang and Xie, Yiming and Jiang, Huaizu},
  journal={arXiv preprint arXiv:2505.19377},
  year={2025}
}

@InProceedings{Guo2022CVPR_humanml3d,
    author    = {Guo, Chuan and Zou, Shihao and Zuo, Xinxin and Wang, Sen and Ji, Wei and Li, Xingyu and Cheng, Li},
    title     = {Generating Diverse and Natural 3D Human Motions From Text},
    booktitle = {Proceedings of the IEEE/CVF Conference on Computer Vision and Pattern Recognition (CVPR)},
    year      = {2022},
}

@inproceedings{zhang2023generating,
    title={T2M-GPT: Generating Human Motion from Textual Descriptions with Discrete Representations},
    author={Zhang, Jianrong and Zhang, Yangsong and Cun, Xiaodong and Huang, Shaoli and Zhang, Yong and Zhao, Hongwei and Lu, Hongtao and Shen, Xi},
    booktitle={Proceedings of the IEEE/CVF Conference on Computer Vision and Pattern Recognition (CVPR)},
    year={2023},
  }

@article{zhang2024motiondiffuse,
  title={Motiondiffuse: Text-driven human motion generation with diffusion model},
  author={Zhang, Mingyuan and Cai, Zhongang and Pan, Liang and Hong, Fangzhou and Guo, Xinying and Yang, Lei and Liu, Ziwei},
  journal={IEEE transactions on pattern analysis and machine intelligence},
  volume={46},
  number={6},
  pages={4115--4128},
  year={2024},
  publisher={IEEE}
}

@article{vaswani2017attention,
  title={Attention is all you need},
  author={Vaswani, Ashish and Shazeer, Noam and Parmar, Niki and Uszkoreit, Jakob and Jones, Llion and Gomez, Aidan N and Kaiser, {\L}ukasz and Polosukhin, Illia},
  journal={Advances in neural information processing systems},
  volume={30},
  year={2017}
}

@article{li2024autoregressive,
  title={Autoregressive Image Generation without Vector Quantization},
  author={Li, Tianhong and Tian, Yonglong and Li, He and Deng, Mingyang and He, Kaiming},
  journal={arXiv preprint arXiv:2406.11838},
  year={2024}
}

@article{chen2024diffusion,
  title={Diffusion forcing: Next-token prediction meets full-sequence diffusion},
  author={Chen, Boyuan and Mart{\'\i} Mons{\'o}, Diego and Du, Yilun and Simchowitz, Max and Tedrake, Russ and Sitzmann, Vincent},
  journal={Advances in Neural Information Processing Systems},
  volume={37},
  pages={24081--24125},
  year={2024}
}

@inproceedings{zhou2019continuity,
  title={On the continuity of rotation representations in neural networks},
  author={Zhou, Yi and Barnes, Connelly and Lu, Jingwan and Yang, Jimei and Li, Hao},
  booktitle={Proceedings of the IEEE/CVF conference on computer vision and pattern recognition},
  pages={5745--5753},
  year={2019}
}

@article{fan2024fluid,
  title={Fluid: Scaling autoregressive text-to-image generative models with continuous tokens},
  author={Fan, Lijie and Li, Tianhong and Qin, Siyang and Li, Yuanzhen and Sun, Chen and Rubinstein, Michael and Sun, Deqing and He, Kaiming and Tian, Yonglong},
  journal={arXiv preprint arXiv:2410.13863},
  year={2024}
}

@misc{rombach2021highresolution,
      title={High-Resolution Image Synthesis with Latent Diffusion Models}, 
      author={Robin Rombach and Andreas Blattmann and Dominik Lorenz and Patrick Esser and Björn Ommer},
      year={2021},
      eprint={2112.10752},
      archivePrefix={arXiv},
      primaryClass={cs.CV}
}

@inproceedings{wang2025motiondreamer,
      title={MotionDreamer: One-to-Many Motion Synthesis with Localized Generative Masked Transformer},
      author={Yilin Wang and chuan guo and Yuxuan Mu and Muhammad Gohar Javed and Xinxin Zuo and Juwei Lu and Hai Jiang and Li cheng},
      booktitle={The Thirteenth International Conference on Learning Representations},
      year={2025},
      url={https://openreview.net/forum?id=d23EVDRJ6g}
      }

@inproceedings{zheng2025autokeyframe,
  title={Autokeyframe: Autoregressive keyframe generation for human motion synthesis and editing},
  author={Zheng, Bowen and Chen, Ke and Yao, Yuxin and Zeng, Zijiao and Jiang, Xinwei and Wang, He and Lasenby, Joan and Jin, Xiaogang},
  booktitle={Proceedings of the Special Interest Group on Computer Graphics and Interactive Techniques Conference Conference Papers},
  pages={1--12},
  year={2025}
}

@inproceedings{pinyoanuntapong2024mmm,
  title={Mmm: Generative masked motion model},
  author={Pinyoanuntapong, Ekkasit and Wang, Pu and Lee, Minwoo and Chen, Chen},
  booktitle={Proceedings of the IEEE/CVF Conference on Computer Vision and Pattern Recognition},
  pages={1546--1555},
  year={2024}
}

@article{EgoLM,
    title={EgoLM: Multi-Modal Language Model of Egocentric Motions},
    author={Fangzhou Hong and Vladimir Guzov and Hyo Jin Kim and Yuting Ye and Richard Newcombe and Ziwei Liu and Lingni Ma},
    journal={arXiv preprint arXiv:2409.18127},
    year={2024}
}

@inproceedings{chen2023executing,
  title={Executing your commands via motion diffusion in latent space},
  author={Chen, Xin and Jiang, Biao and Liu, Wen and Huang, Zilong and Fu, Bin and Chen, Tao and Yu, Gang},
  booktitle={Proceedings of the IEEE/CVF conference on computer vision and pattern recognition},
  pages={18000--18010},
  year={2023}
}

@inproceedings{petrovich2024multi,
  title={Multi-track timeline control for text-driven 3d human motion generation},
  author={Petrovich, Mathis and Litany, Or and Iqbal, Umar and Black, Michael J and Varol, Gul and Bin Peng, Xue and Rempe, Davis},
  booktitle={Proceedings of the IEEE/CVF Conference on Computer Vision and Pattern Recognition},
  pages={1911--1921},
  year={2024}
}

@inproceedings{ahuja2019language2pose,
  title={Language2pose: Natural language grounded pose forecasting},
  author={Ahuja, Chaitanya and Morency, Louis-Philippe},
  booktitle={2019 International conference on 3D vision (3DV)},
  pages={719--728},
  year={2019},
  organization={IEEE}
}

@inproceedings{petrovich2022temos,
  title={Temos: Generating diverse human motions from textual descriptions},
  author={Petrovich, Mathis and Black, Michael J and Varol, G{\"u}l},
  booktitle={European Conference on Computer Vision},
  pages={480--497},
  year={2022},
  organization={Springer}
}

@inproceedings{petrovich2023tmr,
  title={Tmr: Text-to-motion retrieval using contrastive 3d human motion synthesis},
  author={Petrovich, Mathis and Black, Michael J and Varol, G{\"u}l},
  booktitle={Proceedings of the IEEE/CVF International Conference on Computer Vision},
  pages={9488--9497},
  year={2023}
}

@article{baade2026latent,
  title={Latent Forcing: Reordering the Diffusion Trajectory for Pixel-Space Image Generation},
  author={Baade, Alan and Chan, Eric Ryan and Sargent, Kyle and Chen, Changan and Johnson, Justin and Adeli, Ehsan and Fei-Fei, Li},
  journal={arXiv preprint arXiv:2602.11401},
  year={2026}
}

@inproceedings{tevet2022motionclip,
  title={Motionclip: Exposing human motion generation to clip space},
  author={Tevet, Guy and Gordon, Brian and Hertz, Amir and Bermano, Amit H and Cohen-Or, Daniel},
  booktitle={European Conference on Computer Vision},
  pages={358--374},
  year={2022},
  organization={Springer}
}

@inproceedings{kim2023flame,
  title={Flame: Free-form language-based motion synthesis \& editing},
  author={Kim, Jihoon and Kim, Jiseob and Choi, Sungjoon},
  booktitle={Proceedings of the AAAI Conference on Artificial Intelligence},
  volume={37},
  number={7},
  pages={8255--8263},
  year={2023}
}

@inproceedings{he2026molingo,
      title={MoLingo: Motion–Language Alignment for Text-to-Human Motion Generation},
      author={He, Yannan and Tiwari, Garvita and Zhang, Xiaohan and Bora, Pankaj and Birdal, Tolga and Lenssen, Jan Eric and Pons-Moll, Gerard},
      booktitle={CVPR},
      year={2026}
  }

@article{zhang2023remodiffuse,
      title   =   {ReMoDiffuse: Retrieval-Augmented Motion Diffusion Model}, 
      author  =   {Zhang, Mingyuan and
                   Guo, Xinying and
                   Pan, Liang and
                   Cai, Zhongang and
                   Hong, Fangzhou and
                   Li, Huirong and
                   Yang, Lei and
                   Liu, Ziwei},
      year    =   {2023},
      journal =   {arXiv preprint arXiv:2304.01116},
}

@article{chen2025free,
  title={Free-T2M: Frequency Enhanced Text-to-Motion Diffusion Model With Consistency Loss},
  author={Chen, Wenshuo and Jia, Haozhe and Lai, Songning and Wu, Keming and Xiao, Hongru and Hu, Lijie and Yue, Yutao},
  journal={arXiv preprint arXiv:2501.18232},
  year={2025}
}

@article{zhang2023finemogen,
  title={FineMoGen: Fine-Grained Spatio-Temporal Motion Generation and Editing},
  author={Zhang, Mingyuan and Li, Huirong and Cai, Zhongang and Ren, Jiawei and Yang, Lei and Liu, Ziwei},
  journal={NeurIPS},
  year={2023}
}

@article{cho2024discord,
  title={DisCoRD: Discrete Tokens to Continuous Motion via Rectified Flow Decoding},
  author={Cho, Jungbin and Kim, Junwan and Kim, Jisoo and Kim, Minseo and Kang, Mingu and Hong, Sungeun and Oh, Tae-Hyun and Yu, Youngjae},
  journal={arXiv preprint arXiv:2411.19527},
  year={2024}
}

@inproceedings{zhang2025kinmo,
  title={Kinmo: Kinematic-aware human motion understanding and generation},
  author={Zhang, Pengfei and Liu, Pinxin and Garrido, Pablo and Kim, Hyeongwoo and Chaudhuri, Bindita},
  booktitle={Proceedings of the IEEE/CVF International Conference on Computer Vision},
  pages={11187--11197},
  year={2025}
}

@article{liu2025gesturelsm,
  title={GestureLSM: Latent Shortcut based Co-Speech Gesture Generation with Spatial-Temporal Modeling},
  author={Liu, Pinxin and Song, Luchuan and Huang, Junhua and Liu, Haiyang and Xu, Chenliang},
  journal={arXiv preprint arXiv:2501.18898},
  year={2025}
}

@inproceedings{pinyoanuntapong2025maskcontrol,
  title={MaskControl: Spatio-Temporal Control for Masked Motion Synthesis},
  author={Pinyoanuntapong, Ekkasit and Saleem, Muhammad and Karunratanakul, Korrawe and Wang, Pu and Xue, Hongfei and Chen, Chen and Guo, Chuan and Cao, Junli and Ren, Jian and Tulyakov, Sergey},
  booktitle={Proceedings of the IEEE/CVF International Conference on Computer Vision},
  pages={9955--9965},
  year={2025}
}

@article{guo2025snapmogen,
  title={SnapMoGen: Human Motion Generation from Expressive Texts},
  author={Guo, Chuan and Hwang, Inwoo and Wang, Jian and Zhou, Bing},
  journal={arXiv preprint arXiv:2507.09122},
  year={2025}
}

@inproceedings{ghosh2025duetgen,
  title={Duetgen: Music driven two-person dance generation via hierarchical masked modeling},
  author={Ghosh, Anindita and Zhou, Bing and Dabral, Rishabh and Wang, Jian and Golyanik, Vladislav and Theobalt, Christian and Slusallek, Philipp and Guo, Chuan},
  booktitle={Proceedings of the Special Interest Group on Computer Graphics and Interactive Techniques Conference Conference Papers},
  pages={1--11},
  year={2025}
}

@article{liang2024intergen,
  title={Intergen: Diffusion-based multi-human motion generation under complex interactions},
  author={Liang, Han and Zhang, Wenqian and Li, Wenxuan and Yu, Jingyi and Xu, Lan},
  journal={International Journal of Computer Vision},
  pages={1--21},
  year={2024},
  publisher={Springer}
}

@inproceedings{wan2024tlcontrol,
  title={Tlcontrol: Trajectory and language control for human motion synthesis},
  author={Wan, Weilin and Dou, Zhiyang and Komura, Taku and Wang, Wenping and Jayaraman, Dinesh and Liu, Lingjie},
  booktitle={European Conference on Computer Vision},
  pages={37--54},
  year={2024},
  organization={Springer}
}

@article{cai2025flooddiffusion,
  title={FloodDiffusion: Tailored Diffusion Forcing for Streaming Motion Generation},
  author = {Yiyi Cai and Yuhan Wu and Kunhang Li and You Zhou and Bo Zheng and Haiyang Liu},
  journal={arXiv preprint arXiv:2512.03520},
  year={2025}
}

@inproceedings{wewer25srm,
      title     = {Spatial Reasoning with Denoising Models},
      author    = {Wewer, Christopher and Pogodzinski, Bartlomiej and Schiele, Bernt and Lenssen, Jan Eric},
      booktitle = {International Conference on Machine Learning ({ICML})},
      year      = {2025},
}

@article{zhu2025motiongpt3,
  title={MotionGPT3: Human Motion as a Second Modality},
  author={Zhu, Bingfan and Jiang, Biao and Wang, Sunyi and Tang, Shixiang and Chen, Tao and Luo, Linjie and Zheng, Youyi and Chen, Xin},
  journal={arXiv preprint arXiv:2506.24086},
  year={2025}
}

@inproceedings{yuan2023physdiff,
  title={Physdiff: Physics-guided human motion diffusion model},
  author={Yuan, Ye and Song, Jiaming and Iqbal, Umar and Vahdat, Arash and Kautz, Jan},
  booktitle={Proceedings of the IEEE/CVF international conference on computer vision},
  pages={16010--16021},
  year={2023}
}

@InProceedings{dabral2022mofusion,
      title={MoFusion: A Framework for Denoising-Diffusion-based Motion Synthesis},
      author={Rishabh Dabral and Muhammad Hamza Mughal and Vladislav Golyanik and Christian Theobalt},
      booktitle={Computer Vision and Pattern Recognition (CVPR)},
      year={2023}
}

@inproceedings{chen2025language,
  title={The language of motion: Unifying verbal and non-verbal language of 3d human motion},
  author={Chen, Changan and Zhang, Juze and Lakshmikanth, Shrinidhi K and Fang, Yusu and Shao, Ruizhi and Wetzstein, Gordon and Fei-Fei, Li and Adeli, Ehsan},
  booktitle={Proceedings of the Computer Vision and Pattern Recognition Conference},
  pages={6200--6211},
  year={2025}
}

@inproceedings{karunratanakul2023gmd,
  title={Guided Motion Diffusion for Controllable Human Motion Synthesis},
  author={Karunratanakul, Korrawe and Preechakul, Konpat and Suwajanakorn, Supasorn and Tang, Siyu},
  booktitle={Proceedings of the IEEE/CVF International Conference on Computer Vision},
  pages={2151--2162},
  year={2023}
}

@article{zhang2024large,
      title   =   {Large Motion Model for Unified Multi-Modal Motion Generation}, 
      author={Zhang, Mingyuan and Jin, Daisheng and Gu, Chenyang and Hong, Fangzhou and Cai, Zhongang and Huang, Jingfang and Zhang, Chongzhi and Guo, Xinying and Yang, Lei and He, Ying and others},
      year    =   {2024},
      journal =   {arXiv preprint arXiv:2404.01284},
}

@inproceedings{zhangflashmo,
  title={FlashMo: Geometric Interpolants and Frequency-Aware Sparsity for Scalable Efficient Motion Generation},
  author={Zhang, Zeyu and Wang, Yiran and Li, Danning and Gong, Dong and Reid, Ian and Hartley, Richard},
  booktitle={The Thirty-ninth Annual Conference on Neural Information Processing Systems}
}

@inproceedings{tuautoregressive,
  title={Autoregressive Motion Generation with Gaussian Mixture-Guided Latent Sampling},
  author={Tu, Linnan and Meng, Lingwei and Li, Zongyi and Ling, Hefei and Huang, Shijuan},
  booktitle={The Thirty-ninth Annual Conference on Neural Information Processing Systems}
}

@inproceedings{guo2022tm2t,
  title={Tm2t: Stochastic and tokenized modeling for the reciprocal generation of 3d human motions and texts},
  author={Guo, Chuan and Zuo, Xinxin and Wang, Sen and Cheng, Li},
  booktitle={European Conference on Computer Vision},
  pages={580--597},
  year={2022},
  organization={Springer}
}

@article{jiang2023motiongpt,
  title={Motiongpt: Human motion as a foreign language},
  author={Jiang, Biao and Chen, Xin and Liu, Wen and Yu, Jingyi and Yu, Gang and Chen, Tao},
  journal={Advances in Neural Information Processing Systems},
  volume={36},
  pages={20067--20079},
  year={2023}
}

@inproceedings{pinyoanuntapong2024bamm,
  title={Bamm: Bidirectional autoregressive motion model},
  author={Pinyoanuntapong, Ekkasit and Saleem, Muhammad Usama and Wang, Pu and Lee, Minwoo and Das, Srijan and Chen, Chen},
  booktitle={European Conference on Computer Vision},
  pages={172--190},
  year={2024},
  organization={Springer}
}

@inproceedings{zhong2023attt2m,
  title={Attt2m: Text-driven human motion generation with multi-perspective attention mechanism},
  author={Zhong, Chongyang and Hu, Lei and Zhang, Zihao and Xia, Shihong},
  booktitle={Proceedings of the IEEE/CVF international conference on computer vision},
  pages={509--519},
  year={2023}
}

@inproceedings{chen2024taming,
  title={Taming diffusion probabilistic models for character control},
  author={Chen, Rui and Shi, Mingyi and Huang, Shaoli and Tan, Ping and Komura, Taku and Chen, Xuelin},
  booktitle={ACM SIGGRAPH 2024 Conference Papers},
  pages={1--10},
  year={2024}
}

@article{shi2024interactive,
  title={Interactive character control with auto-regressive motion diffusion models},
  author={Shi, Yi and Wang, Jingbo and Jiang, Xuekun and Lin, Bingkun and Dai, Bo and Peng, Xue Bin},
  journal={ACM Transactions on Graphics (TOG)},
  volume={43},
  number={4},
  pages={1--14},
  year={2024},
  publisher={ACM New York, NY, USA}
}

@article{tevet2024closd,
  title={Closd: Closing the loop between simulation and diffusion for multi-task character control},
  author={Tevet, Guy and Raab, Sigal and Cohan, Setareh and Reda, Daniele and Luo, Zhengyi and Peng, Xue Bin and Bermano, Amit H and van de Panne, Michiel},
  journal={arXiv preprint arXiv:2410.03441},
  year={2024}
}

@inproceedings{zhang2024tedi,
  title={Tedi: Temporally-entangled diffusion for long-term motion synthesis},
  author={Zhang, Zihan and Liu, Richard and Hanocka, Rana and Aberman, Kfir},
  booktitle={ACM SIGGRAPH 2024 Conference Papers},
  pages={1--11},
  year={2024}
}

@article{zhao2024dartcontrol,
  title={DartControl: A diffusion-based autoregressive motion model for real-time text-driven motion control},
  author={Zhao, Kaifeng and Li, Gen and Tang, Siyu},
  journal={arXiv preprint arXiv:2410.05260},
  year={2024}
}

@inproceedings{azadi2023make,
  title={Make-an-animation: Large-scale text-conditional 3d human motion generation},
  author={Azadi, Samaneh and Shah, Akbar and Hayes, Thomas and Parikh, Devi and Gupta, Sonal},
  booktitle={Proceedings of the IEEE/CVF International Conference on Computer Vision},
  pages={15039--15048},
  year={2023}
}

@inproceedings{huang2024stablemofusion,
  title={Stablemofusion: Towards robust and efficient diffusion-based motion generation framework},
  author={Huang, Yiheng and Yang, Hui and Luo, Chuanchen and Wang, Yuxi and Xu, Shibiao and Zhang, Zhaoxiang and Zhang, Man and Peng, Junran},
  booktitle={Proceedings of the 32nd ACM International Conference on Multimedia},
  pages={224--232},
  year={2024}
}

@inproceedings{zhang2024motion,
  title={Motion mamba: Efficient and long sequence motion generation},
  author={Zhang, Zeyu and Liu, Akide and Reid, Ian and Hartley, Richard and Zhuang, Bohan and Tang, Hao},
  booktitle={European Conference on Computer Vision},
  pages={265--282},
  year={2024},
  organization={Springer}
}

@inproceedings{zhou2024emdm,
  title={Emdm: Efficient motion diffusion model for fast and high-quality motion generation},
  author={Zhou, Wenyang and Dou, Zhiyang and Cao, Zeyu and Liao, Zhouyingcheng and Wang, Jingbo and Wang, Wenjia and Liu, Yuan and Komura, Taku and Wang, Wenping and Liu, Lingjie},
  booktitle={European Conference on Computer Vision},
  pages={18--38},
  year={2024},
  organization={Springer}
}

@article{lu2023humantomato,
  title={Humantomato: Text-aligned whole-body motion generation},
  author={Lu, Shunlin and Chen, Ling-Hao and Zeng, Ailing and Lin, Jing and Zhang, Ruimao and Zhang, Lei and Shum, Heung-Yeung},
  journal={arXiv preprint arXiv:2310.12978},
  year={2023}
}

@inproceedings{lu2025scamo,
  title={Scamo: Exploring the scaling law in autoregressive motion generation model},
  author={Lu, Shunlin and Wang, Jingbo and Lu, Zeyu and Chen, Ling-Hao and Dai, Wenxun and Dong, Junting and Dou, Zhiyang and Dai, Bo and Zhang, Ruimao},
  booktitle={Proceedings of the Computer Vision and Pattern Recognition Conference},
  pages={27872--27882},
  year={2025}
}

@inproceedings{zhou2024avatargpt,
  title={Avatargpt: All-in-one framework for motion understanding planning generation and beyond},
  author={Zhou, Zixiang and Wan, Yu and Wang, Baoyuan},
  booktitle={Proceedings of the IEEE/CVF Conference on Computer Vision and Pattern Recognition},
  pages={1357--1366},
  year={2024}
}

@inproceedings{fan2025go,
  title={Go to zero: Towards zero-shot motion generation with million-scale data},
  author={Fan, Ke and Lu, Shunlin and Dai, Minyue and Yu, Runyi and Xiao, Lixing and Dou, Zhiyang and Dong, Junting and Ma, Lizhuang and Wang, Jingbo},
  booktitle={Proceedings of the IEEE/CVF International Conference on Computer Vision},
  pages={13336--13348},
  year={2025}
}

@article{fan2024textual,
  title={Textual decomposition then sub-motion-space scattering for open-vocabulary motion generation},
  author={Fan, Ke and Zhang, Jiangning and Yi, Ran and Gong, Jingyu and Wang, Yabiao and Wang, Yating and Tan, Xin and Wang, Chengjie and Ma, Lizhuang},
  journal={arXiv preprint arXiv:2411.04079},
  year={2024}
}

@inproceedings{cen2025ready_to_react,
  title={Ready-to-React: Online Reaction Policy for Two-Character Interaction Generation},
  author={Cen, Zhi and Pi, Huaijin and Peng, Sida and Shuai, Qing and Shen, Yujun and Bao, Hujun and Zhou, Xiaowei and Hu, Ruizhen},
  booktitle={ICLR},
  year={2025}
}

@inproceedings{he2016deep,
  title={Deep residual learning for image recognition},
  author={He, Kaiming and Zhang, Xiangyu and Ren, Shaoqing and Sun, Jian},
  booktitle={Proceedings of the IEEE conference on computer vision and pattern recognition},
  pages={770--778},
  year={2016}
}

@inproceedings{liu2023flow,
  title={Flow Straight and Fast: Learning to Generate and Transfer Data with Rectified Flow},
  author={Xingchao Liu and Chengyue Gong and Qiang Liu},
  booktitle={The Eleventh International Conference on Learning Representations},
  year={2023},
  url={https://openreview.net/forum?id=XVjTT1nw5z}
}

@article{SMPL:2015,
    author = {Loper, Matthew and Mahmood, Naureen and Romero, Javier and Pons-Moll, Gerard and Black, Michael J.},
    title = {{SMPL}: A Skinned Multi-Person Linear Model},
    journal = {ACM Transactions on Graphics, (Proc. SIGGRAPH Asia)},
    month = oct,
    number = {6},
    pages = {248:1--248:16},
    publisher = {ACM},
    volume = {34},
    year = {2015}
}

@inproceedings{mahmood2019amass,
  title={AMASS: Archive of motion capture as surface shapes},
  author={Mahmood, Naureen and Ghorbani, Nima and Troje, Nikolaus F and Pons-Moll, Gerard and Black, Michael J},
  booktitle={Proceedings of the IEEE/CVF international conference on computer vision},
  pages={5442--5451},
  year={2019}
}

@article{ho2022classifier,
  title={Classifier-free diffusion guidance},
  author={Ho, Jonathan and Salimans, Tim},
  journal={arXiv preprint arXiv:2207.12598},
  year={2022}
}

\clearpage
\appendix
In the following, we present additional details and results that complement the main paper. 
We first describe the action plan autoencoder in \cref{subsec:action_ae}, followed by the heterogeneous timestep sampling algorithm in \cref{subsec:timesteps}, architecture and training details in \cref{subsec:arch_details}, and sampling pseudocode in \cref{subsec:sampling_pseudo}. Finally, we present details of our user study in \cref{subsec:user_study_details}.

\section{Action Plan Autoencoder} 
\label{subsec:action_ae}

As described in \cref{subsec:text-align}, we pre-train a lightweight MLP autoencoder to project CLIP text embeddings into a 16-dimensional latent space. This section provides the full architectural and training details.

\vspace{0.1cm} \noindent\textbf{Architecture.}
The encoder and decoder are both 3-layer MLPs with hidden dimensions 256 and 128, BatchNorm and GELU activations. The encoder maps from the CLIP embedding dimension ($d_\text{CLIP} = 512$) to 16, and the decoder maps back from 16 to $d_\text{CLIP}$.

\vspace{0.1cm} \noindent\textbf{Training objectives.}
The autoencoder is trained with three objectives. Given an input CLIP embedding $e \in \mathbb{R}^{d_\text{CLIP}}$, the encoder produces $z = f_\text{enc}(e) \in \mathbb{R}^{16}$ and the decoder reconstructs $\hat{e} = f_\text{dec}(z)$.

\noindent\textit{(1) Reconstruction loss.} We minimize the MSE between the original and reconstructed embeddings in CLIP space:
\begin{equation}
  \mathcal{L}_\text{recon} = \| \hat{e} - e \|^2_2.
\end{equation}

\noindent\textit{(2) Neighbor-preservation loss.} To preserve semantic similarity in the latent space, we encourage correct action label retrieval. For each input $e \in \mathbb{R}^{d_\text{CLIP}}$, we pre-compute its ground-truth label $y$ as the index of the nearest neighbor of $e$ in a fixed label bank $L \in \mathbb{R}^{6133 \times d_\text{CLIP}}$. We then treat retrieval as classification: for the reconstructed embedding $\hat{e}$, we compute cosine similarity scores $s_i = (\hat{e}^\top \ell_i) / (\tau \| \hat{e} \| \| \ell_i \|)$ against each normalized label $\ell_i$, and minimize cross-entropy against $y$: 

\begin{equation}
  \mathcal{L}_\text{neighbor} = -\log \frac{\exp(s_y / \tau)}{\sum_{j=1}^{K} \exp(s_j / \tau)} = \text{CE}\left( \text{softmax}\left( \frac{\hat{e}^\top L^\top}{\tau \| \hat{e} \| \| L \|} \right), y \right),
\end{equation}
where $L$ is the $6133 \times d_\text{CLIP}$ label bank (rows $\ell_i$), $y$ is the ground-truth nearest-neighbor index, and $\tau$ is a temperature hyperparameter (default $\tau = 0.07$).

\noindent\textit{(3) Variance regularization.} To prevent dimensional collapse, we encourage each latent dimension to maintain unit variance across the training batch:
\begin{equation}
  \mathcal{L}_\text{var} = \sum_{j=1}^{16} \left( \text{Var}(z_j) - 1 \right)^2,
\end{equation}
where $z_j$ denotes the $j$-th dimension across all training samples in a batch.
The total autoencoder loss is:
\begin{equation}
  \mathcal{L}_\text{AE} = \mathcal{L}_\text{recon} + \lambda_\text{n} \, \mathcal{L}_\text{neighbor} + \lambda_\text{v} \, \mathcal{L}_\text{var}.
\end{equation}
We use $\lambda_\text{n} = 1.0$ and $\lambda_\text{v} = 0.01$ by default.

\noindent\textbf{Text retrieval at inference.} Following~\cite{li2025unimotion}, we decode the 16-d latents back to CLIP space using the decoder and retrieve the nearest action label via KNN over the CLIP embeddings of all action labels in the BABEL vocabulary.

\section{Heterogeneous Timestep Sampling}
\label{subsec:timesteps}
As discussed in \cref{subsec:training}, each motion latent $x_i$ is assigned an independent timestep $t_i^x$ during training. Naively sampling each $t_i^x \sim \mathcal{U}(0,1)$ independently would cause the sequence mean $\bar{t}^x = \frac{1}{N}\sum_i t_i^x$ to concentrate around $0.5$ due to the Bates distribution, undertraining the model at early ($\bar{t} \approx 0$) and late ($\bar{t} \approx 1$) denoising stages~\cite{chen2024diffusion}.

Following SRM~\cite{wewer25srm}, we instead first sample a mean timestep $\bar{t}^x \sim \mathcal{U}(0,1)$ and then perturb individual timesteps around this mean:
\begin{equation}
  t_i^x = \mathrm{clip}\!\left(\bar{t}^x + \delta_i,\; 0,\; 1\right),
  \quad \delta_i \sim \mathcal{N}(0, \sigma_t^2),
\end{equation}
where $\sigma_t$ controls the spread of individual timesteps. This ensures that $\bar{t}^x$ remains uniformly distributed across training, providing balanced coverage of all denoising stages. We set $\sigma_t = \text{1.0}$ in all experiments.

\section{Architecture and Training Details}
\label{subsec:arch_details}

\vspace{0.1cm} \noindent\textbf{Transformer denoiser architecture.}
As stated in the main paper, our denoiser $G_\theta$ is a Transformer with 16 layers, 16 attention heads, and a hidden dimension of 1024. At each temporal position, the motion latent $x_i$ and action plan latent $y_i$ are concatenated into a single vector and linearly projected to the hidden dimension. 

\vspace{0.1cm} \noindent\textbf{Training hyperparameters.}
We train with the Adam optimizer with a learning rate of 1e-4 and batch size of 32 on 4 NVIDIA A100 GPUs for 10,000 epochs, which takes approximately 26 hours. The loss weights are set to $\lambda_x = 1$ and $\lambda_y = 1$.
We apply classifier-free guidance~\cite{ho2022classifier} by randomly dropping the sequence-level text condition $c$ with probability $0.1$ during training and using a guidance scale of $5.5$ at inference.

\vspace{0.1cm} \noindent\textbf{Dataset details.}
We utilize both HumanML3D~\cite{Guo2022CVPR_humanml3d} and BABEL~\cite{BABEL:CVPR:2021}, which independently annotate different subsets of AMASS~\cite{mahmood2019amass}. The full HumanML3D-272~\cite{xiao2025motionstreamer} training set contains 21,466 motion sequences with sequence-level text descriptions. Of these, only 8,829 sequences overlap with BABEL and have additional frame-level action labels. We do not apply left-right flipping augmentation when computing this overlap statistic.

\section{Sampling Pseudocode}
\label{subsec:sampling_pseudo}
We provide detailed pseudocode for both offline (\cref{alg:offline}) and streaming inference (\cref{alg:streaming}) modes. Both share the same Stage~1 (action plan generation) but differ in how motion latents are denoised in Stage~2.

\begin{algorithm}[h]
\caption{Offline Mode Sampling}
\begin{algorithmic}[1]
\Require Text condition $c$, denoising steps $T$, overlap steps $K$, sequence length $N$
\Ensure Clean motion latents $\hat{\mathbf{x}}_0$
\State \textbf{Stage 1: Action plan generation}
\State Sample $\mathbf{x} \sim \mathcal{N}(0, \mathbf{I}), \mathbf{y} \sim \mathcal{N}(0, \mathbf{I})$, $t_i^x = 1 $ for $\forall i$ \Comment{Initialization}
\For{$s = 0$ to $T-1$}
    \State $t^y = 1 - s / T$
    \State $(\_, \hat{\mathbf{y}}_v) = G_\theta(\mathbf{x}_1, \mathbf{y}; \mathbf{1}, t^y, c)$ \Comment \cf \cref{eq:denoiser},$\mathbf{t}_x = \mathbf{1}$ and $\mathbf{x}_{\mathbf{1}}$ are fixed for Stage 1
    \State $\mathbf{y} \leftarrow \mathbf{y} - \hat{\mathbf{y}}_v / T$ \Comment Update $\mathbf{y}$ with the predicted velocity
\EndFor
\State $\hat{\mathbf{y}}_0 \leftarrow \mathbf{y}$ \Comment{The clean action plan}
\State
\State \textbf{Stage 2: Progressive motion denoising in random order}
\State Generate random permutation $\pi$ of $\{1, \dots, N\}$
\State Initialize active set $\mathcal{A} = \emptyset$, per-latent progress $p_i = 0$ for $\forall i$, step counter $S=0$
\While{any $p_i < T$}
    \If{$\pi \neq \emptyset$ and $S \bmod K = 0$} \Comment Activate the next latent every $K$ steps
        \State $\mathcal{A} \leftarrow \mathcal{A} \cup \{\pi_k\}$ \Comment Activate the next random latent $\pi_k$
        \State $\pi \leftarrow \pi \setminus \{\pi_k\}$ \Comment Remove
    \EndIf
    \For{each $i \in \mathcal{A}$ with $p_i < T$} \Comment Active latents \textbf{NOT} fully denoised
        \State $t_i^x = 1 - p_i / T$ \Comment Update timesteps
    \EndFor
    \State $(\hat{\mathbf{x}}_v, \_) = G_\theta(\mathbf{x}, \hat{\mathbf{y}}_{0}; \mathbf{t}^x, \mathbf{0}, c)$ \Comment \cf \cref{eq:denoiser}, action plan $\hat{\mathbf{y}}_0$ is fixed for Stage 2
    \For{each $i \in \mathcal{A}$ with $p_i < T$}
        \State $x_i = x_i - \hat{x}_{v,i}/T$ \Comment Update the active motion latents only
        \State $p_i \leftarrow p_i + 1$
    \EndFor
    \State $S \leftarrow S+1$
\EndWhile
\State $\hat{\mathbf{x}}_0 \leftarrow \mathbf{x}$
\end{algorithmic}
\label{alg:offline}
\end{algorithm}

\begin{algorithm}[h]
\caption{Streaming Mode Sampling}
\begin{algorithmic}[1]
\Require Text condition $c$, denoising steps $T$, latent length $N \ (T < N)$
\Ensure Streamed clean motion frames $\hat{\mathbf{x}}_0$
\State \textbf{Stage 1: Action plan generation with the first motion latent fully denoised}
\State Sample $\mathbf{x} \sim \mathcal{N}(0, \mathbf{I}), \mathbf{y} \sim \mathcal{N}(0, \mathbf{I})$ \Comment Initialization
\State Initialize active set $\mathcal{A} = \emptyset$, per-motion latent progress $p_i = 0$ for $\forall i$
\For{$s = 0$ to $T-1$}
    \State $t^y = 1 - s/T$
    \State $\mathcal{A} \leftarrow \mathcal{A} \cup \{x_s\}$ \Comment Activate the next latent $x_s$
    \For{each $i \in \mathcal{A}$ with $p_i < T$} \Comment Active latents \textbf{NOT} fully denoised
        \State $t_i^x = 1 - p_i / T$ \Comment Update timesteps
    \EndFor
    \State $(\hat{\mathbf{x}}_v, \hat{\mathbf{y}}_v) = G_\theta(\mathbf{x}, \mathbf{y}; \mathbf{t}^x, t^y, c)$ \Comment \cf \cref{eq:denoiser}
    \For{each $i \in \mathcal{A}$ with $p_i < T$}
        \State $x_i = x_i - \hat{x}_{v,i}/T$ \Comment Update the active motion latents only
        \State $p_i \leftarrow p_i + 1$
    \EndFor
    \State $\mathbf{y} \leftarrow \mathbf{y} - \hat{\mathbf{y}}_v / T$ \Comment Update the action plan
\EndFor
\State $\hat{\mathbf{y}}_0 \leftarrow \mathbf{y}$ \Comment{The clean action plan}
\State $s \leftarrow s+1$
\State Decode $x_1$ with the Causal TAE decoder to obtain motion frames
\State
\State \textbf{Stage 2: Sequential progressive denoising in raster order}
\While{any $p_i < T$}
    \If{$s < N$}
        \State $\mathcal{A} \leftarrow \mathcal{A} \cup \{x_s\}$ \Comment Activate the next latent $x_s$
    \EndIf
    \For{each $i \in \mathcal{A}$ with $p_i < T$} \Comment Active latents \textbf{NOT} fully denoised
        \State $t_i^x = 1 - p_i / T$ \Comment Update timesteps
    \EndFor
    \State $(\hat{\mathbf{x}}_v, \_) = G_\theta(\mathbf{x}, \hat{\mathbf{y}}_{0}; \mathbf{t}^x, \mathbf{0}, c)$ \Comment \cf \cref{eq:denoiser}, action plan is fixed
    \For{each $i \in \mathcal{A}$ with $p_i < T$}
        \State $x_i = x_i - \hat{x}_{v,i}/T$ \Comment Update the active motion latents only
        \State $p_i \leftarrow p_i + 1$
        \If{$p_i = T$}
            \State Decode $x_i$ with the Causal TAE decoder to obtain motion frames
        \EndIf
    \EndFor
    \State $s \leftarrow s + 1$
\EndWhile
\State $\hat{\mathbf{x}}_0 \leftarrow \mathbf{x}$
\end{algorithmic}
\label{alg:streaming}
\end{algorithm}

\section{User Study Details}
\label{subsec:user_study_details}

\begin{figure}[h]
  \centering
  \includegraphics[width=\linewidth]{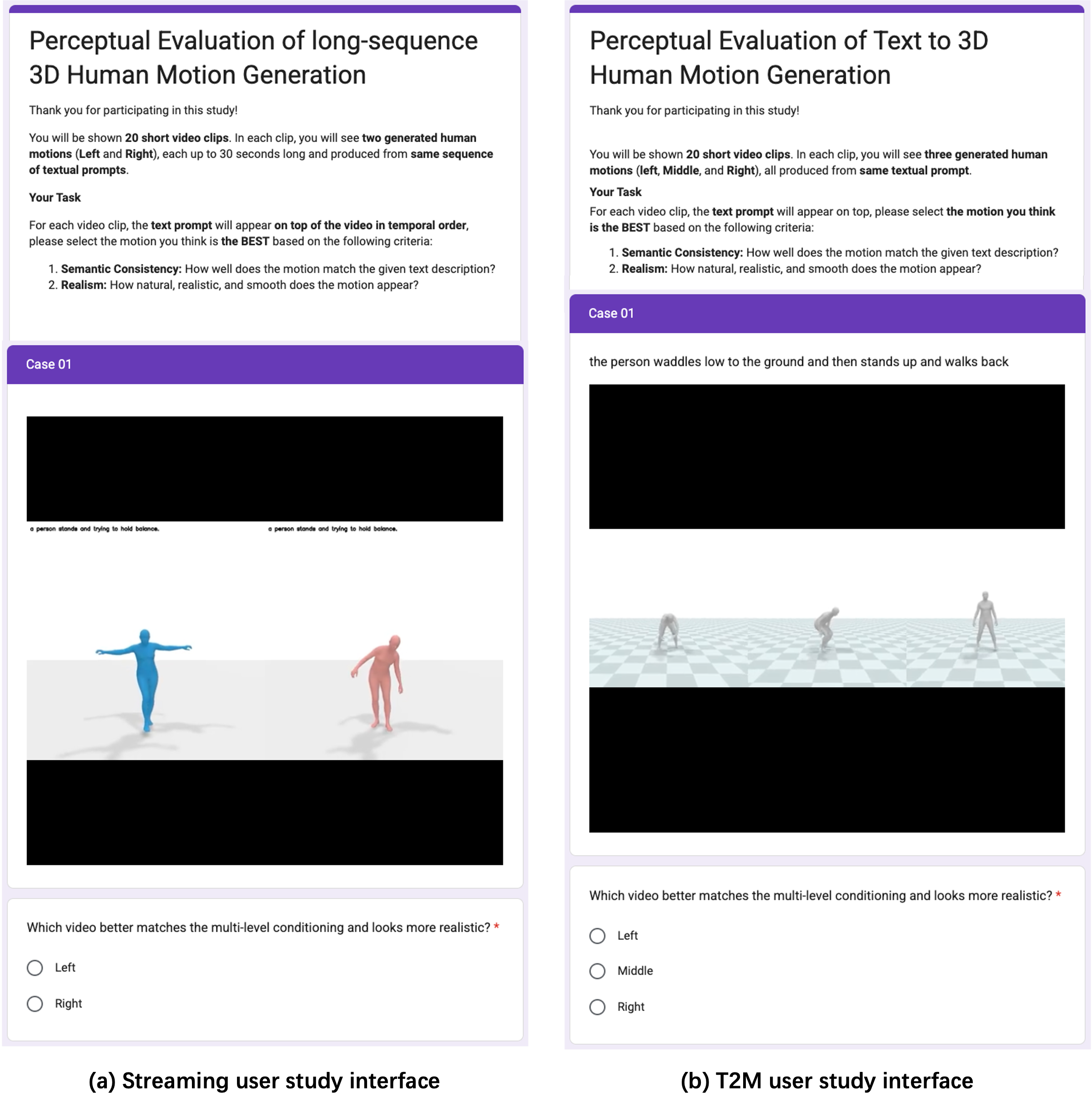}
  \caption{\textbf{User study interfaces.} (a) Long-sequence streaming evaluation: participants compare two animations (left vs.\ right) generated from the same sequence of textual prompts. Note that the left side is always shown in blue and the right side is always shown in red, while the actual method assigned to each side is randomized. (b) Text-to-motion evaluation: participants compare three animations (left, middle, right) generated from the same single prompt. In both studies, method assignment is randomized and participants select the best motion based on semantic consistency and realism.}
  \label{fig:user_study}
\end{figure}

We provide additional details on the user study interface and design. The detailed instructions and interface layouts are shown in \cref{fig:user_study}.

\vspace{0.1cm} \noindent\textbf{Text-to-motion generation study (\cref{fig:user_study}b).}
Each trial presents three side-by-side animations produced from the same textual prompt, with method assignment to left, middle, and right positions randomized. The text prompt is displayed on top of the video. Participants are asked to select the motion that best matches the text description and looks most realistic.

\vspace{0.1cm} \noindent\textbf{Long motion generation study (\cref{fig:user_study}a).}
Each trial presents two side-by-side animations generated from the same sequence of textual prompts (up to 30 seconds), with left-right assignment randomized. The text prompts are displayed on top of the video in temporal order so that participants can verify whether each action is faithfully executed. Participants select the preferred motion using the same two criteria.

\end{document}